%% file: main.tex
\PassOptionsToPackage{table}{xcolor}
\documentclass{pretty-preprint}

\usepackage[section]{placeins}

\usepackage{amsthm,amsmath,amssymb}

\usepackage{graphicx}
\usepackage{subcaption}
\usepackage{tikz}

\usepackage{booktabs}
\usepackage{multirow}
\usepackage{tabularx}
\usepackage{array}
\usepackage{nicematrix}

\usepackage{xcolor}
\usepackage{caption}
\usepackage[most]{tcolorbox}

\usepackage[numbers,square,sort&compress]{natbib}
\usepackage{subcaption}
\usepackage{capt-of}

\newsavebox{\RAASpaceTableBox}

\definecolor{takeawayBg}{HTML}{F7F2FC}
\definecolor{takeawayBorder}{HTML}{D8C8F7}

\definecolor{ArenaSection}{HTML}{F2F2F2}

\title{%
  \includegraphics[width=0.5\textwidth]{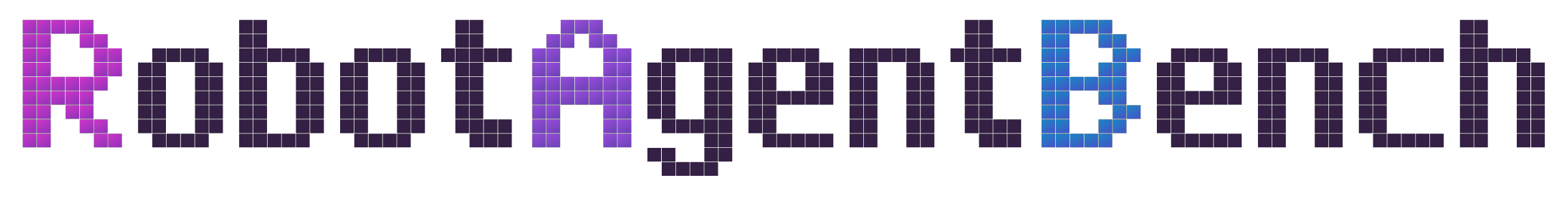}\\[-2pt]
  {\LARGE Towards Open-world Robot Autonomy}%
}
\author[1,2,†,‡]{Lingjun Mao}
\author[1,2,†,‡]{Jinglin Cao}
\author[1,2,†]{Lukun He}
\author[1,2]{Wenpeng Xu}
\author[1,2]{Yuchen Yan}
\author[1]{Hengyu Jin}
\author[1]{Zijun Zhang}
\author[1,2]{Jinzhou Tang}
\author[1,2]{Songyao Jin}
\author[2]{Ziche Liu}
\author[1]{Yifei Shao}
\author[1]{Junbo Huang}
\author[1,2]{Fang Nan}
\author[1,2]{Ruobing Han}
\author[1]{Ziqiao Xi}
\author[1]{Ziming Xu}
\author[1,2]{Shuang Liang}
\author[1,2]{Sibo Zhu}
\author[1,2]{Wenyi Wu}
\author[1,2]{Xinyue Wang}
\author[1,2]{Yiming Huang}
\author[1,2]{Xinkai Zou}
\author[3]{Gao Zhu}
\author[1,*,‡]{Kun Zhou}
\author[1]{Biwei Huang}

\affiliation[1]{Aether AI}
\affiliation[2]{University of California San Diego}
\affiliation[3]{University of California Davis}

\contribution[†]{Equal contribution}
\contribution[‡]{Co-project leaders}
\contribution[*]{Corresponding author}

\date{\today}
\correspondence{Kun Zhou (\email{franciskunzhou@gmail.com})}
\metadata[Code]{\href{https://github.com/AetherLabsAI/RobotAgentBench}{https://github.com/AetherLabsAI/RobotAgentBench}}
\metadata[Model]{\href{https://huggingface.co/AetherLabs-AI/InternVLA-A1.5-RAB}{https://huggingface.co/AetherLabs-AI/InternVLA-A1.5-RAB}}
\metadata[Project]{\href{https://aetherlabsai.github.io/RobotAgentBench/}{https://aetherlabsai.github.io/RobotAgentBench/}}

\input{sections/abstract}


\begin{document}
\maketitle

\input{sections/introduction}

\input{sections/robot_agent_arena}

\input{sections/experiments}

\input{sections/related_works}

\section{Conclusion}
\label{sec}

In this work, we introduce \textit{RobotAgentBench} (RAB), a comprehensive benchmark for evaluating open-world robot autonomy beyond predefined manipulation tasks. Built upon an agent-assisted real-to-sim pipeline, RAB provides multi-level evaluations spanning five fundamental capabilities and open-ended household scenarios, enabling systematic assessment of robot agents from task discovery and contextual reasoning to physical execution and adaptive recovery. Through extensive evaluations of VLA/WAM models and diverse robot agent architectures, we reveal a substantial gap between manipulation capabilities and autonomous task completion. Our findings demonstrate the benefits of coding-agent harnesses, hybrid execution, and recursive self-improvement, while highlighting persistent challenges in planning–execution coordination, failure recovery, and real-world deployment. Despite recent advances, reliable autonomy in open-ended scenarios remains far from solved. 

We hope RAB serves as a systematic testbed for understanding these limitations and advancing the transition from robot manipulation toward open-world robot autonomy. Building on RAB, we aim to extend agent-level recursive self-improvement (RSI) into the physical open world. Through autonomous exploration and interaction, robot agents could discover new tasks, learn from failures, and continuously refine their planning and execution strategies. Accumulating and transferring reusable experience across tasks and environments could pave the way toward continually self-improving robot autonomy with minimal human intervention.

\bibliography{paper}
\bibliographystyle{authordate1}

% Appendix draft; task and training details remain subject to author review.
\clearpage
\appendix
\begingroup
\fontsize{10}{12}\selectfont
\input{appendix/appendix}
\endgroup

\end{document}

%% file: sections/abstract.tex
\abstract{
Recent advances in vision-language-action (VLA) and world-action models (WAMs) have significantly improved robotic manipulation, yet deploying robots in the open world demands a higher level of autonomy beyond executing predefined tasks. Robot agents offer a promising paradigm for this transition, integrating task discovery, contextual reasoning, physical execution, and adaptive recovery into a closed interaction loop. However, existing benchmarks largely evaluate isolated capabilities, failing to capture the complexity of open-world robot autonomy. We introduce RobotAgentBench (RAB), a comprehensive benchmark for evaluating robot agents in realistic, open-world household environments. Through an agent-assisted, human-in-the-loop real-to-sim pipeline, RAB constructs 50 diverse household layouts with interactive objects and physically grounded tasks, encompassing 90 atomic tasks, 18 short-horizon tasks, 8 long-horizon tasks, and 4 open-ended scenarios. RAB provides a multi-level evaluation of robot agent capabilities: (1) task-level assessments systematically examine five complementary dimensions—task discovery, context reasoning, manipulation, failure recovery, and robustness; (2) scenario-level evaluation assesses how agents integrate these capabilities to autonomously fulfill high-level user needs. Furthermore, RAB enables systematic comparisons of standalone VLA/WAM models and diverse robot agent architectures under a unified planning–execution–verification framework. Extensive experiments reveal substantial gaps between manipulation and autonomy, demonstrate the advantages of agentic architectures and recursive self-improvement, and highlight the remaining challenges toward open-world robot autonomy.

}

%% file: sections/introduction.tex
\section{Introduction}
\label{intro}
Embodied AI has advanced rapidly in recent years, with vision-language-action (VLA) and world-action models (WAMs) demonstrating increasingly strong manipulation capabilities across diverse tasks and environments~\citep{intelligence2025pi_, ye2026world, team2026xiaomi, wu2026foundation, cai2026internvla, xu2026causalwm, shao2026causalwam}. These advances raise a fundamental question: \emph{How far are we from deploying robots in the open world?} Unlike predefined manipulation tasks, the real world involves numerous mundane yet complex challenges, often characterized by ambiguity, uncertainty, and constantly changing circumstances~\citep{liu2025robopilot, pchelintsev2025lera, gong2026replan}. Consider a household robot preparing for guests: it must identify what needs to be cleaned, select appropriate food based on user preferences, coordinate multiple activities, and recover when manipulation attempts fail~\citep{mao2026cris0}. Such challenges are not exceptional, but intrinsic to everyday physical environments. Addressing them is essential for transforming robots into broadly useful assistants in human environments.

Operating in the open world demands a transition \emph{from robot manipulation to robot autonomy}. Beyond executing explicit instructions, robots must autonomously discover relevant tasks, reason about environmental and user context, plan physical interactions, and adapt to unexpected outcomes. Robot agents provide a natural framework for this transition, integrating high-level reasoning and planning with physical execution and feedback-driven verification into a closed loop~\citep{team2026systematically, shi2025hi, ahn2022can, huang2022inner}. Such agents must determine not only how to execute, but also what to do and how to adapt. Meanwhile, advances in multimodal foundation models, including GPT-6-Astra, and emerging agentic embodied systems~\citep{team2025gemini, yang2025agentic, bjorck2026vesta, chen2026eta} are creating new opportunities to realize this integration. Together, these developments suggest that embodied AI is entering a new stage, shifting from standalone manipulation models toward autonomous robot agents capable of handling complex, dynamic physical environments.
%These capabilities must operate in a closed loop, enabling robots to determine not only how to execute, but also what to do and how to adapt. Meanwhile, advances in multimodal foundation models, including GPT-6-Astra, and emerging agentic embodied systems~\citep{team2025gemini, yang2025agentic, bjorck2026vesta, chen2026eta} are opening new opportunities to integrate high-level reasoning with physical execution. Together, these developments suggest that embodied AI is entering a new stage where autonomous decision-making, rather than manipulation alone, becomes a central challenge.

However, existing benchmarks remain largely designed for robot manipulation rather than autonomy. Widely adopted benchmarks~\citep{rlbench, metaworld, maniskill2, libero, robocasa, mu2024robotwin, robotwin2} primarily evaluate predefined tasks with explicit goals, overlooking autonomous task discovery and context-dependent decision-making. Recent efforts extend evaluation toward reasoning~\citep{chen2026robodojo} and long-horizon planning~\citep{alfred, alfworld, realfred, embodiedbench, mao2025deliverybench}, but often rely on stylized problems or abstracted physical interactions. More importantly, reasoning, planning, and manipulation are frequently evaluated in isolation, without capturing their interdependence during real-world operation. A robot capable of individual tasks may still struggle to determine what needs to be done, recover from failures, or adapt to human interventions. Evaluating open-world robot autonomy therefore requires moving beyond isolated capabilities toward a comprehensive assessment of the entire autonomous interaction loop.

To bridge this gap, we introduce RobotAgentBench~(RAB), a benchmark for evaluating open-world robot autonomy in realistic household environments. We develop an agent-assisted, human-in-the-loop real-to-sim pipeline to reconstruct real-world scenes and expand them into 50 diverse layouts with interactive objects and physically grounded tasks. 
At its core, RAB provides 90 atomic tasks that can be flexibly composed into a wide range of household tasks, from which we construct 18 short-horizon and 8 long-horizon manipulation tasks as the foundation of our evaluation. Building upon these tasks, we introduce diverse variants and evaluation settings to systematically assess five complementary dimensions of robot autonomy: (1) task discovery, identifying what needs to be done; (2) context reasoning, interpreting implicit requirements; (3) manipulation, executing physical actions; (4) failure recovery, detecting and correcting execution errors; and (5) robustness, adapting to environmental disturbances. Beyond these task-level assessments, we introduce four open-ended household scenarios that require robot agents to integrate these capabilities to autonomously fulfill high-level user needs, enabling an end-to-end evaluation of open-world robot autonomy.
%Our benchmark is built upon 90 atomic actions, composed into 18 short-horizon and 8 long-horizon household manipulation tasks. Building on this task suite, we introduce diverse task variants and evaluation settings to assess five complementary evaluation dimensions, alongside four open-ended household scenarios: 
%Our benchmark comprises 90 atomic tasks, 18 short-horizon tasks, 8 long-horizon composite tasks, and four open-ended household scenarios. To systematically characterize robot autonomy, we introduce five complementary evaluation dimensions: 
%(1) task discovery, identifying what needs to be done; (2) context reasoning, interpreting implicit requirements; (3) manipulation, executing physical actions; (4) failure recovery, detecting and correcting execution errors; and (5) robustness, adapting to environmental disturbances. Beyond these task-level assessments, our scenario-level evaluation measures whether robot agents can integrate these capabilities to autonomously fulfill high-level user needs, providing an end-to-end assessment of robot autonomy.

Using RobotAgentBench, we systematically evaluate standalone VLA/WAM models and diverse robot agent architectures, revealing five key findings: 
\begin{itemize}
    \setlength{\itemsep}{0pt}
    \setlength{\parsep}{0pt}
    \item Standalone policies struggle with task discovery and contextual reasoning despite their manipulation capabilities, highlighting the gap between manipulation and autonomy; 
    \item Coding-agent harnesses substantially improve performance, particularly through hybrid code-based and learned-policy execution; 
    \item Both planner capabilities and executor instruction-following abilities critically influence agent performance;
    \item Harness-level recursive self-improvement (RSI) progressively improves task- and scenario-level performance through reusable tool refinement and experience accumulation;
    \item Real-world evaluations reveal performance degradation but broadly consistent relative rankings, supporting the benchmark's relevance to physical deployment. 
\end{itemize}
Together, these findings highlight the promise and remaining challenges of advancing from robot manipulation toward open-world robot autonomy.

%% file: sections/robot_agent_arena.tex
\begin{figure*}[t]
    \centering
    \includegraphics[width=\linewidth]{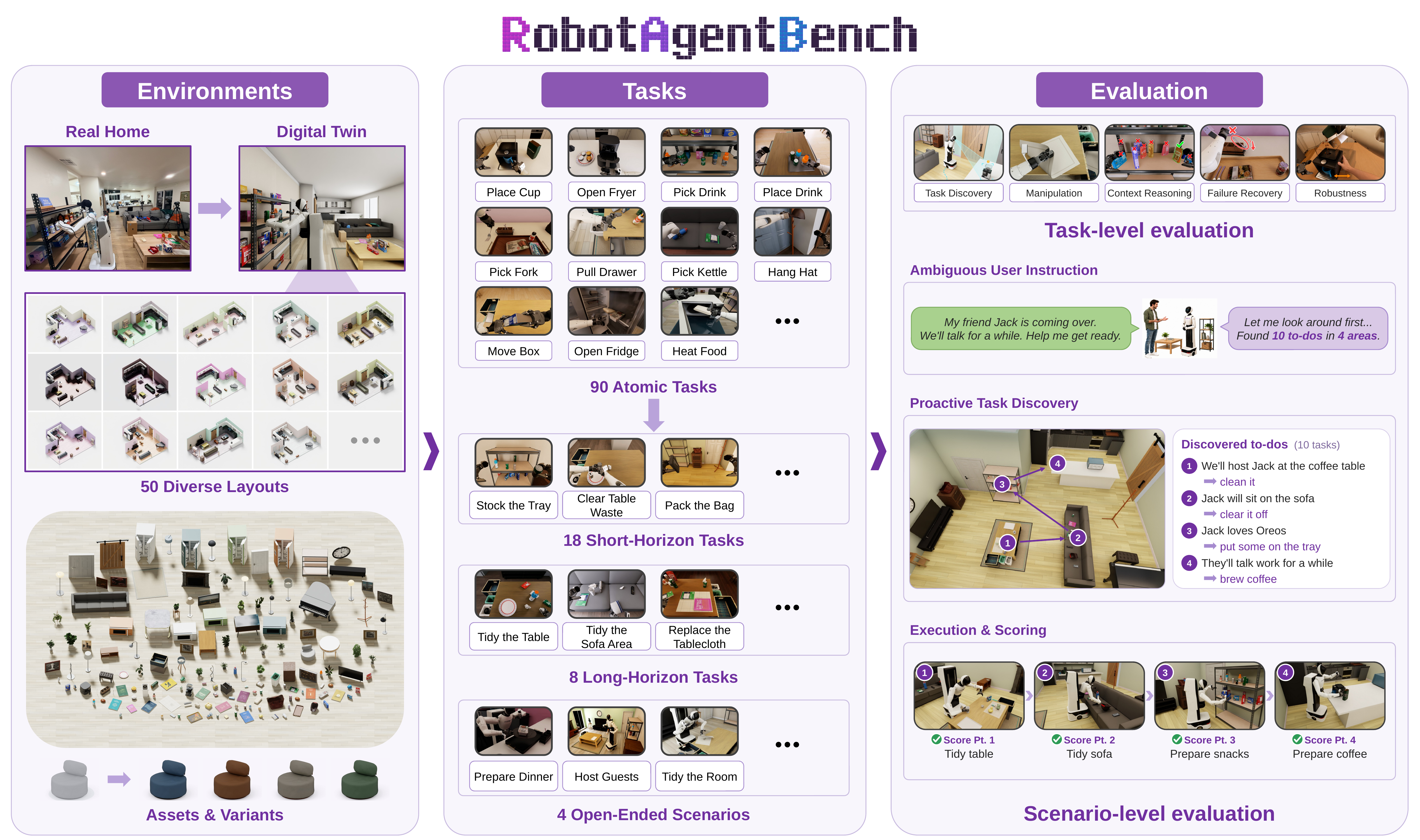}
    \caption{Overview of RobotAgentBench.}
    \label{fig:RobotAgentBench}
\end{figure*}

\section{RobotAgentBench}
\label{sec:robot_agent_arena}

%We present \textit{RobotAgentBench}, a benchmark for evaluating robots' planning, reasoning, and manipulation capabilities at both the task and scenario levels. We develop a real-to-sim pipeline for environment construction and data collection, alongside a unified framework that organizes representative robot agent designs for systematic comparison.
We introduce \textit{RobotAgentBench}, a benchmark designed to evaluate open-world robot autonomy beyond predefined manipulation tasks. We leverage a real-to-sim pipeline to construct diverse environments and collect data, then build the multi-level evaluation suite for systematic comparisons of diverse agent framework designs. Figure~\ref{fig:RobotAgentBench} illustrates the overview of our benchmark.

\subsection{Real-to-Sim Pipeline for Environment and Data Construction}
%To ground the tasks in realistic household interactions, we reconstruct a real home in Isaac Sim and efficiently expand it into diverse layouts and furniture configurations. We then use physical-robot demonstrations as references for generating training data in simulation (Fig.~\ref{fig:RobotAgentArena}).
%To support realistic and scalable evaluation, we develop an agent-assisted, human-in-the-loop real-to-sim pipeline that reconstructs a physical environment based on a real  household scene. Then, we expand it into diverse interactive settings and collect demonstrations for training and test.
To enable realistic and scalable evaluation, we develop an agent-assisted, human-in-the-loop real-to-sim pipeline that starts from a physical room in a real home, reconstructs it in simulation, expands the reconstructed scene into diverse interactive configurations, and collects demonstrations for training and evaluation.

% \begin{figure}[!htbp]
%     \centering
%     \includegraphics[width=\linewidth]{figure_a_real2sim_pipeline.pdf}
%     \caption{Human-in-the-loop Real2Sim. Environment construction conditions action generation. Physics rollouts expose interaction gaps, and human review guides refinements to assets, physical parameters or actions before re-execution.}
%     \label{fig:real2sim-pipeline}
% \end{figure}

\subsubsection{Environment Construction}
Starting from a physical room in a real home, we first reconstruct it in Isaac Sim and then expand it into diverse household environments. We subsequently collect real-robot demonstrations and reproduce them in simulation to generate trajectories.

\paragraph{Household Scene Reconstruction.}
%We reconstruct a real home in Isaac Sim using photographs and measurements of its floor plan, furniture arrangement, and household objects. A coding agent retrieves similar 3D assets online and iteratively adjusts their geometry, scale, and appearance to match these references. For simple objects such as snack packages, it directly generates geometry and textures from the photographs. The agent then arranges the assets according to the real-world layout and assigns collision geometry and physical properties to each object. We manually inspect each asset for visual fidelity and physical plausibility, providing feedback to guide further refinement where needed.
We reconstruct the physical household environment in Isaac Sim through an iterative pipeline comprising real-world parameter collection, similar asset retrieval, and new asset reconstruction. First, we collect photographs and measurements of the room layout, furniture arrangements, and household objects to establish real-world references. Next, a coding agent retrieves similar 3D assets from online repositories and adjusts their geometry, scale, and appearance to match these references. For objects without suitable existing assets, we leverage the agent to directly reconstruct their geometry and textures from photographs. The resulting assets are assembled into the simulated environment with appropriate collision geometry and physical properties to support realistic interactions. Throughout this iterative process, human reviewers continuously validate visual fidelity and physical interaction consistency against the real-world environment, providing feedback to the agent for refining the reconstructed scene.
%We first reconstruct a real home in Isaac Sim using photographs and measurements of its floor plan, furniture arrangements, and household objects. A coding agent retrieves suitable 3D assets online and iteratively adjusts their geometry, scale, and appearance to match the real-world references. For simple objects such as snack packages, the agent directly generates geometry and textures from photographs. It then assembles the scene and assigns collision geometry and physical properties to enable realistic interactions. Throughout this process, human reviewers inspect visual fidelity and physical plausibility, providing feedback for iterative refinement.

% \begin{figure}[!htbp]
%     \centering
%     \includegraphics[width=0.72\linewidth]{figure_b_real2sim_assets.pdf}
%     \caption{Representative Real2Sim asset reconstructions. Real reference photographs are cropped and glare-reduced; simulation views use approximately matched viewpoints and object scales.}
%     \label{fig:real2sim-assets}
% \end{figure}

%\begin{samepage}
\paragraph{Environment Augmentation.}
%Starting from the reconstructed home, we create additional layouts by modifying the overall floor plan (e.g., T-shaped and L-shaped configurations), rearranging functional areas, and varying furniture positions and orientations. Each layout is manually reviewed to eliminate geometric errors, such as furniture interpenetration, and implausible arrangements, such as a sofa facing away from the television. This process yields 50 distinct layouts with substantial spatial variation. We further increase visual diversity by varying furniture colors, textures, materials, and interior design styles.
Building upon the reconstructed home, we systematically diversify its spatial configurations and visual appearances. We generate 50 distinct layouts by modifying floor plans (\emph{e.g.,} T-shaped and L-shaped configurations), rearranging functional areas, and varying furniture positions and orientations. Each layout is manually reviewed to eliminate geometric inconsistencies and implausible arrangements. We further increase environmental diversity through variations in furniture colors, textures, materials, and interior design styles.

%\par
%\end{samepage}

\subsubsection{Demonstration Collection}
%We adopt the AgiBot G1 as the robot embodiment, matching its camera configurations and joint settings between simulation and the physical platform. We first collect teleoperated demonstrations in the real home and replay them in simulation, iteratively refining asset geometry and physical parameters to ensure consistent interaction outcomes. Building on these demonstrations, we employ a coding agent to generate additional trajectories across diverse object positions, environment layouts, and initial robot configurations. With access to the full simulator state, including object poses, the agent synthesizes and executes action programs to accomplish target tasks. For challenging interactions, we provide manually implemented control functions that the agent can invoke and compose. This process enables scalable data generation while maintaining consistency with real-world demonstrations. Detailed observation and action spaces are provided in Appendix~\ref{app:benchmark_setup}.
We follow a three-stage workflow comprising real-world demonstration collection, simulation alignment, and agent-assisted trajectory expansion to generate scalable, physically grounded data. First, we adopt the AgiBot G1 as the robot embodiment, matching its camera configurations and joint settings between simulation and the physical platform, and collect teleoperated demonstrations in the real home. Second, we replay these demonstrations in simulation, iteratively refining asset geometry and physical parameters to align interaction outcomes with the real world. Finally, building upon these reference trajectories, we employ a coding agent to generate additional successful trajectories across diverse object positions, environment layouts, and initial robot configurations. With access to the full simulator state, including object poses, the agent synthesizes and executes action programs to accomplish target tasks. For challenging interactions, we provide manually implemented control functions that the agent can invoke and compose. This workflow enables scalable trajectory generation while maintaining consistency with real-world demonstrations. Detailed observation and action spaces are provided in Appendix~\ref{app:benchmark_setup}.

%We use the AgiBot G1 as the robot embodiment, with camera views and joint configurations matched between simulation and the physical platform. Detailed observation, state, and action spaces are provided in Appendix~\ref{app:benchmark_setup}. Using this matched setup, we collect demonstrations through teleoperation in the real home and replay them in simulation, iteratively refining asset geometry and physical parameters to obtain consistent interaction outcomes. We then generate additional training trajectories under variations in target object positions, environment layouts, and initial robot configurations. Guided by the replayed demonstrations and given access to the full simulator state, including the 3D poses of all objects, a coding agent synthesizes and executes action programs to produce successful trajectories. For complex tasks where autonomous program synthesis has a low success rate, we provide manually written rule-based functions for the agent to invoke and compose.
%Figure~\ref{fig:real-openended-case-study} illustrates real-world task-level robustness and scenario-level open-ended execution through two rows of chronologically ordered snapshots.

\subsection{Multi-level Evaluation of Robot Autonomy}
Open-world robot autonomy requires robots to determine what needs to be done, execute physical tasks, and adapt to unexpected outcomes. To systematically evaluate these capabilities, RAB adopts a multi-level evaluation suite built upon a composable household task foundation. At the task level, we introduce dedicated evaluation settings to diagnose five complementary capabilities: task discovery, context reasoning, manipulation, failure recovery, and robustness. At the scenario level, we assess whether robot agents can integrate these capabilities to autonomously fulfill high-level user needs in open-ended household environments. Together, these two levels enable both fine-grained capability assessment and end-to-end evaluation of robot autonomy.

%Our benchmark spans a broad range of task complexity, from individual manipulation tasks to open-ended, long-horizon household scenarios. All tasks and scenarios are grounded in real-world household activities. At the task level, we assess five key capabilities: manipulation, task discovery, context reasoning, failure recovery, and robustness. At the scenario level, we evaluate whether robot agents can flexibly integrate these capabilities to complete tasks and meet users' needs in realistic, open-ended household settings.

\begin{figure*}[t]
    \centering
    \includegraphics[width=\linewidth]{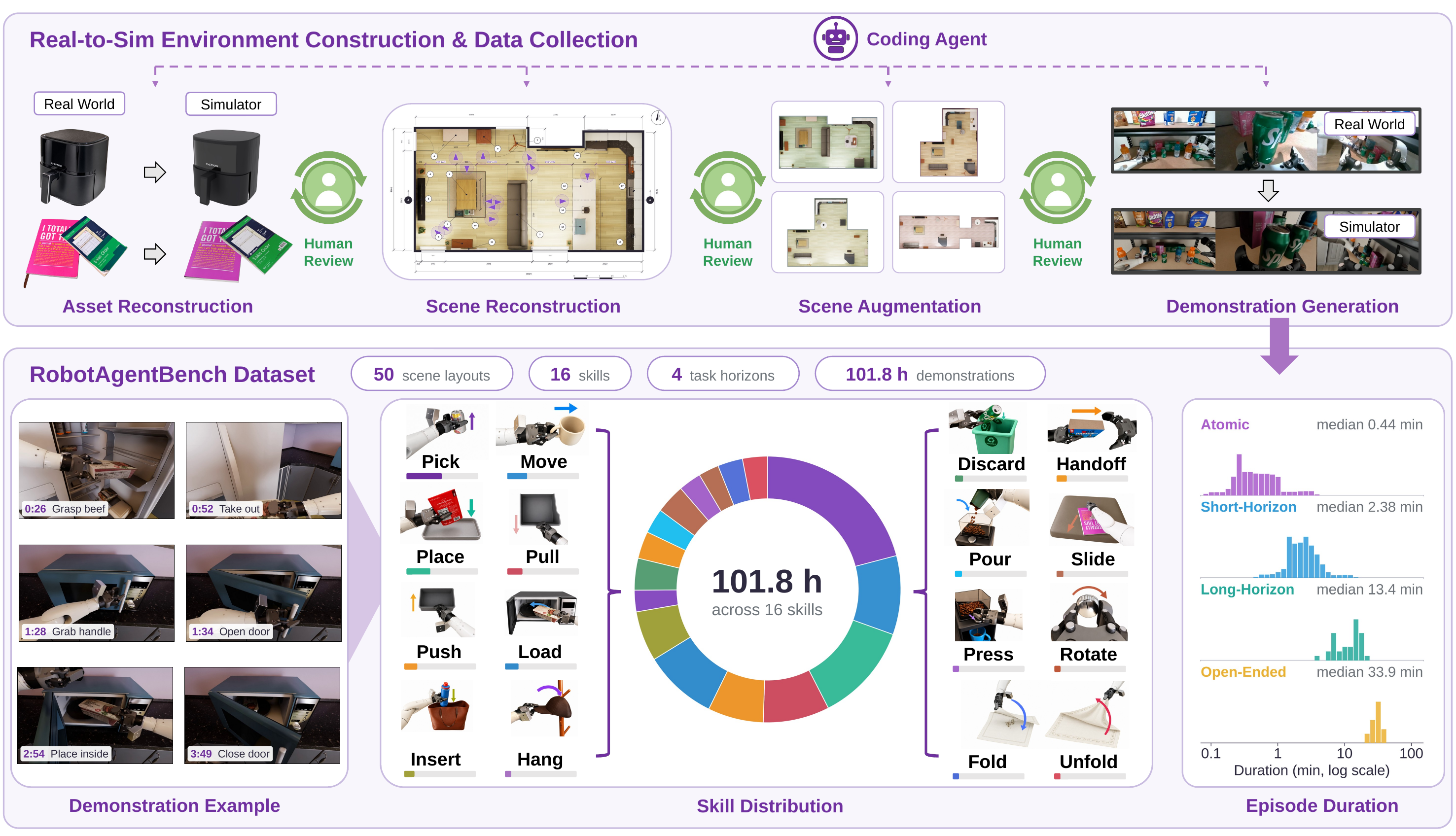}
    \caption{Real-to-sim pipeline and training data statistics for RobotAgentBench.}
    \label{fig:DataPipeline}
\end{figure*}

\subsubsection{Composable Household Task Suite}
We construct a foundational task suite comprising 90 atomic tasks involving everyday household objects and appliances. These tasks can be flexibly composed into diverse household tasks, from which we construct 18 representative short-horizon tasks and 8 long-horizon composite tasks. The suite covers common activities, such as retrieving food from a refrigerator, heating meals in a microwave, and using an air fryer. These tasks involve diverse manipulation skills, including pushing, pulling, bimanual coordination, and mobile manipulation, across rigid objects, deformable materials (\emph{e.g.,} tablecloths), and granular materials (\emph{e.g.,} coffee beans). Short-horizon tasks combine a few atomic tasks, such as picking up a hat from a sofa and hanging it on a coat rack, while long-horizon tasks involve multiple subgoals, such as tidying an entire sofa area. Building upon this task suite, we construct diverse evaluation settings by varying task instructions, contextual requirements, initial states, and environmental disturbances. These variants enable targeted assessments of different autonomy capabilities without requiring entirely new manipulation tasks. We further compose these tasks into open-ended household scenarios, where the tasks to be performed depend on the environment and user context.
For each atomic task and short-horizon task, we collect 50 training trajectories; for each long-horizon task and open-ended scenario, we provide 10 complete demonstrations. In total, the dataset comprises 101.8 hours of training data. Further details are provided in
Figure~\ref{fig:DataPipeline}.

%\subsubsection{Basic Task Suite}
%Our basic task suite comprises 90 diverse atomic actions involving everyday household objects and appliances. We combine these actions into 18 representative short-horizon tasks and 8 long-horizon composite tasks. The suite covers common household activities, such as retrieving food from a refrigerator, heating meals in a microwave, and using an air fryer. These activities require a range of skills, including pushing, pulling, bimanual coordination, and mobile manipulation, across rigid bodies, deformable objects (e.g., tablecloths), and granular materials (e.g., coffee beans). Short-horizon tasks combine a few atomic actions (e.g., picking up a hat from a sofa and hanging it on a coat rack), while long-horizon tasks involve multiple subgoals (e.g., tidying the entire sofa area).

%These tasks and their variants (e.g., tasks perturbed by disturbances during execution) form the basis of our task-level evaluation. Further combinations of these tasks define four open-ended household scenarios, with task selection guided by the environment and user context.

\subsubsection{Task-level Capability Evaluation}
We design five complementary evaluation settings to examine how robots determine what to do, execute physical actions, and adapt to unexpected outcomes. Each setting focuses on a specific capability through controlled task configurations. We use capability-specific metrics: target-object coverage for task discovery, target-selection accuracy for context reasoning, and task progress scores for manipulation, failure recovery, and robustness.

\paragraph{Task Discovery.}
We evaluate whether robots can autonomously identify relevant tasks without explicit task instructions. We configure household scenes with task-relevant objects and provide scenario descriptions and user-specific context. The robot must explore the environment, interpret its observations, and infer which objects require attention. To reduce manipulation demands, the robot indicates inferred targets by touching the relevant objects with either gripper, without completing the underlying household tasks. We measure the fraction of target objects touched under a limited contact budget.

\paragraph{Context Reasoning.}
We assess whether robots can infer appropriate actions from implicit requirements rather than explicit object instructions. We construct four task settings involving drink selection, snack selection, waste selection, and sofa object selection. Given contextual information such as user preferences, intended uses, or classification rules, the robot must identify the appropriate target object. For example, waste selection requires identifying the object that belongs to the requested category under the supplied classification rules. We evaluate reasoning through contact with the correct target object, without requiring successful grasping or manipulation.

\paragraph{Manipulation.}
We evaluate physical execution through the 18 short-horizon and 8 long-horizon tasks in the foundational task suite. Given explicit task instructions, robots must execute the required physical interactions, assessing their instruction-following and manipulation capabilities across diverse objects, skills, and task horizons.

\paragraph{Failure Recovery.}
We evaluate whether robots can recognize and recover from execution failures. We construct five failure variants of the foundational tasks, including missed grasps and objects accidentally knocked over during manipulation. Starting from these failure states, robots must inspect the environment, identify the failure, and execute corrective actions to resume and complete the specified manipulation objective.

\paragraph{Robustness.}
We assess whether robots can adapt to unexpected environmental changes during execution. We construct six task variants involving human interventions and lighting changes, such as moving a coffee machine while the robot is pouring beans into it. Robots must respond to these disturbances and adjust their actions to complete the specified manipulation objective.

\subsubsection{Scenario-level Integrated Evaluation}
Beyond individual capabilities, open-world robot autonomy requires their integration into a closed loop of task discovery, reasoning, planning, execution, and adaptation. We therefore introduce four open-ended household scenarios: \textit{preparing dinner},
\textit{hosting guests}, \textit{preparing to leave},
and \textit{tidying the room}. Each scenario involves context-dependent combinations of tasks, with specific requirements determined by the sampled environment and user information. Unlike task-level evaluation, robots do not receive explicit task lists. Instead, they must autonomously explore the environment, identify relevant tasks, interpret user needs, and plan their execution. This includes reasoning about task dependencies, such as retrieving food from the refrigerator before heating it in the microwave, and coordinating multiple physical interactions toward broader objectives. 

We evaluate end-to-end autonomy using the final progress score
aggregated over all task checkpoints in each scenario.
To further analyze policy and agent behavior, we introduce four
detailed metrics informed by the task-level capability dimensions:
\textit{Task Coverage}, the fraction of scenario tasks with recorded
task-relevant contact or native operation-start evidence;
\textit{Task Progress}, the total process score earned across
these engaged tasks divided by their total maximum process score;
\textit{Recovery Rate}, the fraction of observable physical-failure
segments followed by success of the same task; and
\textit{Stagnation}, the fraction of execution time spent in repetitive,
unproductive attempts or states with no task progress.
Appendix~\ref{app:evaluation_reporting} gives the operational definitions
and aggregation rules.

\subsection{Robot Agent Framework for Evaluation}
\label{sec:agent_arc}

To systematically investigate the transition from standalone manipulation models to autonomous robot agents, RobotAgentBench supports the evaluation of both standalone VLA/WAM policies and diverse robot agent architectures. We introduce a unified planning-execution-verification framework for robot agents: (1) a \emph{planner} that interprets environmental context and user requirements, decomposes tasks, and adapts plans based on execution feedback; (2) an \emph{executor} that translates instructions into physical actions through learned policy models (\emph{e.g.,} VLA and WAM) or code-based control; and (3) a \emph{verifier} that monitors execution outcomes, identifies failures, and guides corrective actions. By flexibly combining these components, we construct five representative agent architectures to systematically examine the roles of planning, execution, verification, and recursive self-improvement in robot autonomy. For clarity, we denote each architecture by its planner/executor/verifier (P/E/V) configuration, where C represents code-based control, L denotes a learned VLA/WAM policy, H denotes hybrid code-based and learned execution, and – indicates an absent component.

%Robot Agent Arena supports evaluation of both standalone policy models and integrated robot agents. Standalone policies include VLAs and WAMs, which generate robot actions conditioned on observations and task instructions. To systematically compare robot agent designs, we introduce a lightweight modular harness organized around a planner, an executor, and a verifier. The planner interprets scenario context and ambiguous user requests, decomposes tasks into concrete instructions, and replans based on execution outcomes. The executor translates these instructions into physical actions, while the verifier checks execution and guides corrections or retries when problems arise. Different harness designs instantiate, combine, or omit these components and vary how they interact within a unified planning--execution--verification framework.

\paragraph{VLA/WAM-based Policy + Code-based Verifier (–/L/C).}
A VLA/WAM policy directly generates actions from observations and task instructions. A coding-agent verifier monitors execution feedback, selectively executes action prefixes, and introduces local corrections before querying the policy again~\citep{team2026systematically}.

%Without an explicit planner, the learned policy directly receives observations and the original task instruction or contextual prompt. A coding-agent verifier reviews policy-generated actions and execution feedback, executing action prefixes or introducing local corrective motions before querying the policy again~\citep{team2026systematically}.

\paragraph{Code-based Planner + VLA/WAM-based Policy (C/L/–).}
A coding-agent planner interprets task requirements, decomposes them into concrete subtask instructions, and delegates physical execution to a VLA/WAM policy. The planner dynamically revises its instructions based on execution feedback~\citep{shi2025hi, ahn2022can, huang2022inner}.

%A coding-agent planner issues concrete subtask instructions to a learned policy executor and revises them based on execution feedback~\citep{shi2025hi, ahn2022can, huang2022inner}.

\paragraph{Code-based Planner + Code-based Policy + Code-based Verifier (C/C/C).}
A coding agent jointly handles planning, execution, and verification by generating executable programs from predefined control functions (\emph{e.g.,} \texttt{move\_eef\_to()}). It composes these functions to accomplish tasks and verifies execution outcomes to guide subsequent actions~\citep{liang2023code, huang2023instruct2act, huang2023voxposer}.

%A coding agent handles all three roles, generating and executing programs composed of simple, manually written control functions (e.g., \texttt{move\_eef\_to()}) and checking their outcomes~\citep{liang2023code, huang2023instruct2act, huang2023voxposer}.

\paragraph{Code-based Planner + Hybrid Code/VLA/WAM-based Policy + Code-based Verifier (C/H/C).}
A coding agent serves as both planner and verifier, while the executor integrates code-based control and learned VLA/WAM policies as unified callable tools. The agent dynamically selects and combines these tools according to task requirements and environmental feedback, leveraging their complementary capabilities.

%A coding agent serves as both planner and verifier, treating code-based control and learned policies as uniformly callable execution tools and selecting between them based on the current subtask and observations.

\paragraph{Code-based Planner + Code-based Policy + Code-based Verifier + RSI (C/C/C + RSI).}
Building upon the fully code-based architecture, we introduce harness-level recursive self-improvement (RSI), enabling the coding agent to iteratively create, refine, and reuse control tools based on execution feedback. Successful improvements are retained across subsequent attempts, supporting progressive adaptation and experience accumulation~\citep{physicalrsi2026}.
%This configuration extends the fully code-based harness with recursive self-improvement (RSI), allowing the coding agent to create and modify control tools based on execution feedback~\citep{physicalrsi2026}. Effective revisions are retained across attempts.

%% file: sections/experiments.tex
\section{Experiments}
\label{results}

We conduct systematic experiments at both the task and scenario levels to evaluate standalone VLA/WAM-based policies and complete robot agents. Building on the unified robot agent framework we developed, we compare different harness designs and examine their effects on task performance. We further evaluate agents in the corresponding real-world environment to investigate performance gaps between simulation and reality.

\subsection{Experimental Setup}

\paragraph{Evaluated Models and Agents.} For standalone policy baselines, we evaluate VLA models and WAMs.
The VLA baselines include $\pi_{0.5}$~\citep{intelligence2025pi_}, Xiaomi Robotics-1~\citep{team2026xiaomi},
InternVLA-A1.5~\citep{ma2026internvla}, MolmoAct2~\citep{fang2026molmoact2}, LingBot-VLA2~\citep{wu2026foundationapplicationimprovingvla}, HY-VLA~\citep{zhang2026hy},
Galaxea G0.5~\citep{liu2026g0}, and X-VLA~\citep{zheng2026x}.
The WAM baselines include $\tau_0$ (low-level policy)~\citep{zhou2026tau0wmunifiedvideoactionworld}, OpenWAM-Alpha~\citep{wang2026openwam}, and LingBot-VA~\citep{li2026causal}. For complete robot agents, we compare five representative
harness designs, using Codex powered by GPT-6 Astra~\citep{openai2026astra} as the coding agent throughout and InternVLA-A1.5 as the learned policy wherever
applicable. For the fully code-based design, we additionally evaluate Claude Code powered by Opus 5.5~\citep{anthropic2026opus55} and Antigravity CLI powered by Gemini 4 Argon~\citep{google2026argon} to compare different coding agents under the same harness. For the RSI setting, we run ten iterations, evaluating 30 instances per task or scenario in each iteration, and report performance in the final iteration.

\paragraph{Evaluation Protocol.}
We evaluate all methods across the five task-level assessments and four open-ended scenarios. 
For each task or scenario, we evaluate 30 distinct instances across the full benchmark suite. Evaluation covers all four open-ended scenario families and the 50-layout catalog, with every applicable randomization axis enabled over its full benchmark-defined range: layout, appearance, lighting, scene and physical distractors, and robot-arm, mobile-base, and interaction-object initial states. The resulting comparisons measure performance across randomized instances within the benchmark scene, task, and context families. All methods interact with the same environment and 22-dimensional command interface; model-specific observation adapters are described in Appendice~\ref{app:policy_training}. For open-ended scenarios, episodes terminate if the environment's evaluator records no increase in task progress for 120 consecutive seconds of simulation time. Coding agents are limited to 200 model responses per episode, and each tool call executes at most 600 control steps, corresponding to 30 seconds of simulation time. We report the mean score under the corresponding evaluation metric for each task or scenario. Comparisons and rank orders refer to these observed instance means under the stated protocol. 

\paragraph{Observation and Action Spaces.}
We adopt observation and action spaces consistent with the physical AgiBot G1 robot. All systems have access to the same observation sources: RGB images from one head-mounted and two wrist-mounted cameras, proprioceptive states, task instructions or contextual prompts, and a progressively revealed navigation map derived from static scene geometry under shared disclosure rules. Model inputs follow their native adapters; LingBot-VA conditions on RGB and language without proprioceptive state. The 22-dimensional action space controls the head, torso, both arms and grippers, and mobile base. Detailed specifications are provided in Appendix A.

\subsection{Main Results}

Table~\ref{tab:main_results} summarizes the performance of standalone policy models and robot agents under both task-level and scenario-level evaluation.

\input{tables/table1}

\subsubsection{Task-Level Evaluation}
We compare standalone VLA/WAM-based policies and complete robot agent systems across the five capability dimensions. Standalone policies generally underperform integrated agent systems, highlighting a substantial gap between their current capabilities and the demands of real-world autonomy.

\paragraph{Standalone Policy Results.}
Standalone policies appear to primarily imitate demonstrated manipulation behaviors, with limited generalization to situations beyond the training distribution, such as recovering from execution failures. They struggle particularly with task discovery, which requires purposeful exploration rather than execution of a specified task. When initialized away from viewpoints commonly encountered in the demonstrations, these models often produce erratic movements instead of navigating to gather relevant observations. They also struggle with contextual reasoning, such as selecting a beverage that matches user preferences, with performance generally close to random guessing. Among the evaluated standalone models, InternVLA-A1.5 achieves the strongest overall performance, with notably higher robustness and failure recovery scores than other models, even surpassing some complete robot agents.

\paragraph{Robot Agent Results.}
Introducing coding agents substantially improves robot performance.
Among fully code-based systems, Antigravity CLI has the highest reported
mean score, followed by Claude Code.
For Codex, RSI further improves performance
through tool refinement and experience accumulation across task iterations. We also compare three ways of integrating coding agents with learned policies. Refining policy-generated actions (-/L/C) yields modest gains but remains constrained by the underlying policy's capabilities. Using a coding agent as a planner (C/L/-) improves manipulation and
robustness, with the largest gains in context reasoning, as the planner
translates contextual information into explicit instructions for the executor.
The best performance comes from combining code-based control with
selectively invoked learned policies (C/H/C). For example, code-based
control positions the gripper near an object, providing a favorable
starting state for the policy to complete the grasp.

\begin{figure}[!t]
    \centering
    \includegraphics[width=\linewidth]{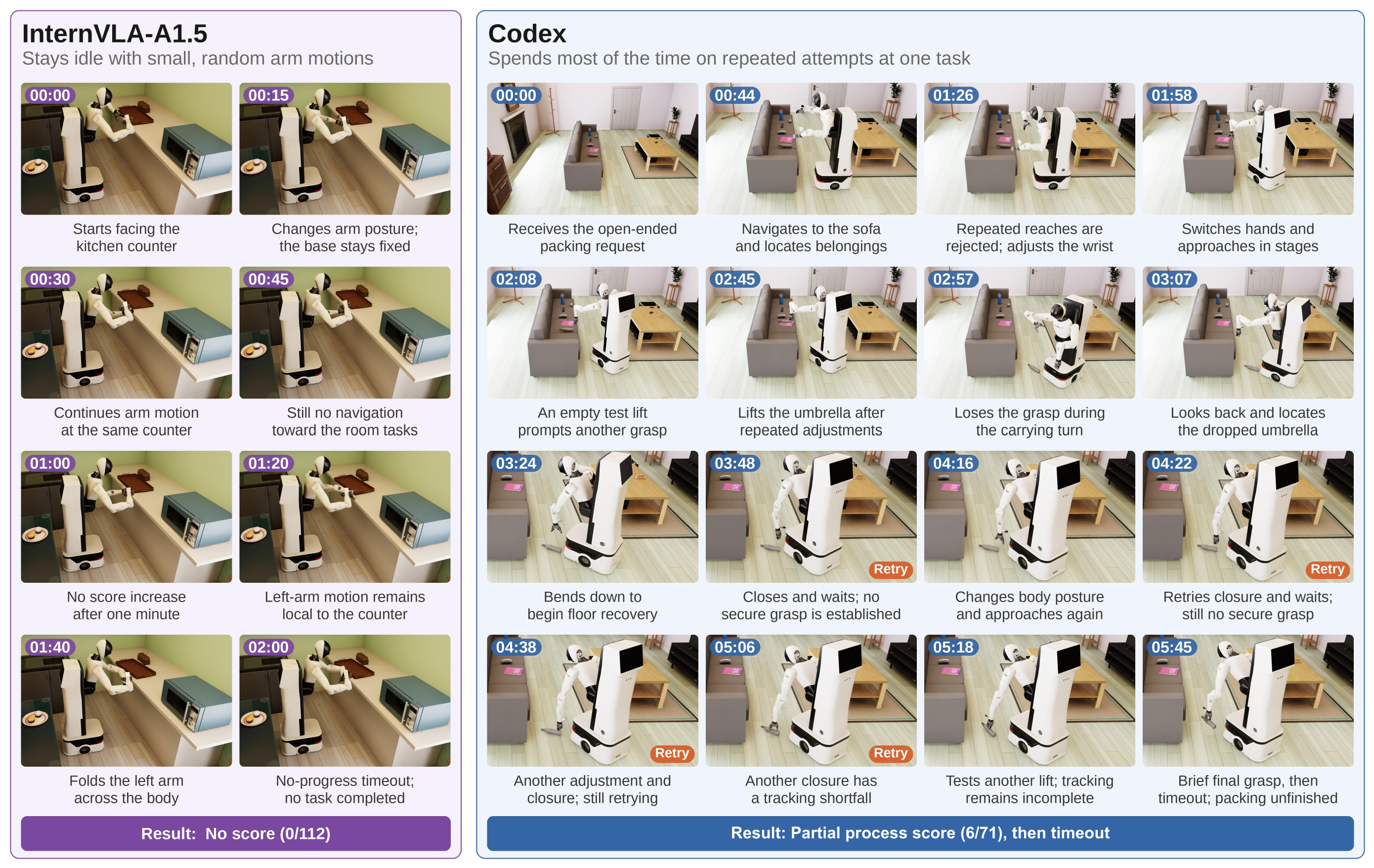}
    \caption{Open-ended scenario case studies. InternVLA-A1.5 (left) fails to initiate any task, while Codex (right) repeatedly attempts a single subtask without completing it before the episode times out.}
    \label{fig:openend_case_study}
\end{figure}

\begin{table*}[!t]
    \centering
    \caption{Detailed evaluation on open-ended scenarios.
    Coverage, Progress, and Recovery denote Task Coverage,
    Task Progress, and Recovery Rate, respectively.
    $\uparrow$/$\downarrow$ indicate higher/lower is better.}
    \label{tab:open_ended_analysis}

    \input{tables/table3}
\end{table*}

\subsubsection{Scenario-Level Evaluation}
Both standalone policies and robot agents perform poorly in open-ended
scenarios. Table~\ref{tab:open_ended_analysis} provides a more detailed
behavioral analysis. Due to their limited task discovery capabilities,
VLA/WAM-based policies rarely explore their surroundings in response
to contextual information. They spend most of their time idle or
making random, unproductive movements, with occasional progress
primarily when initialized near a relevant interaction site.
Complete robot agents identify tasks more actively, and their planners
often generate appropriate task instructions. However, frequent
execution errors leave them stuck on individual tasks, repeatedly
retrying and adjusting their actions. Figure~\ref{fig:openend_case_study} illustrates these behaviors
through representative case studies.

% \begin{tcolorbox}[
%     title={Takeaway 1: Strong instruction following makes $\pi_{0.5}$ a more effective executor for open-ended agents.},
%     colback=takeawayBg,
%     colframe=takeawayBorder,
%     coltitle=black
% ]
% While advanced VLM planners achieve similar planning performance, agent success varies substantially across executors. $\pi_{0.5}$ follows planner-generated instructions more reliably and achieves higher execution accuracy, translating strong high-level plans into substantially better end-to-end performance.
% \end{tcolorbox}

\subsection{Agent Harness Analysis}
Building on our evaluation of robot agent systems, we further investigate how different harness designs affect robot performance, including which coding agents are most effective as planners and which policy models integrate best with coding-agent harnesses. We also examine how agent performance evolves over multiple rounds of recursive self-improvement.

\begin{figure*}[!h]
    \centering
    \begin{subfigure}{0.95\linewidth}
        \centering
        \includegraphics[width=\linewidth]{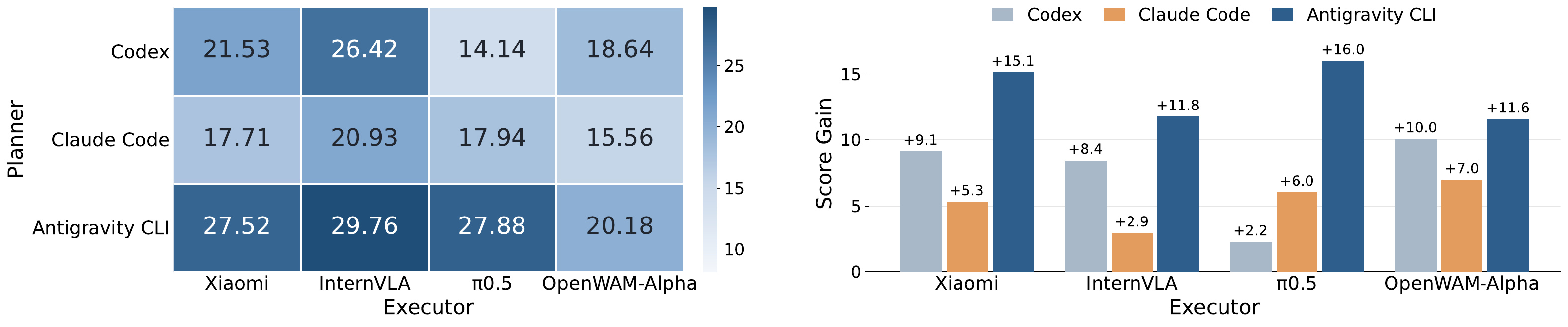}
        \caption{Task-level performance across different planner--executor combinations.}
        \label{fig:task_level}
    \end{subfigure}

    \vspace{0.5em}

    \begin{subfigure}{0.95\linewidth}
        \centering
        \includegraphics[width=\linewidth]{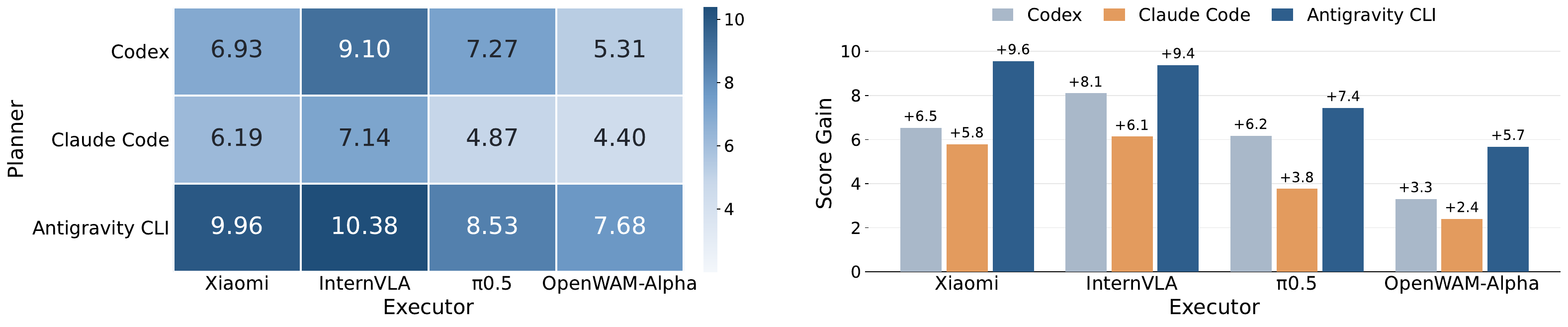}
        \caption{Scenario-level performance across different planner--executor combinations.}
        \label{fig:open_ended}
    \end{subfigure}

    \caption{Performance comparison of planner--executor combinations on RobotAgentBench at the task level and in open-ended scenarios.}
    \label{fig:agent_performance}
\end{figure*}

\begin{figure*}[!h]
    \centering
    \includegraphics[width=0.9\linewidth]{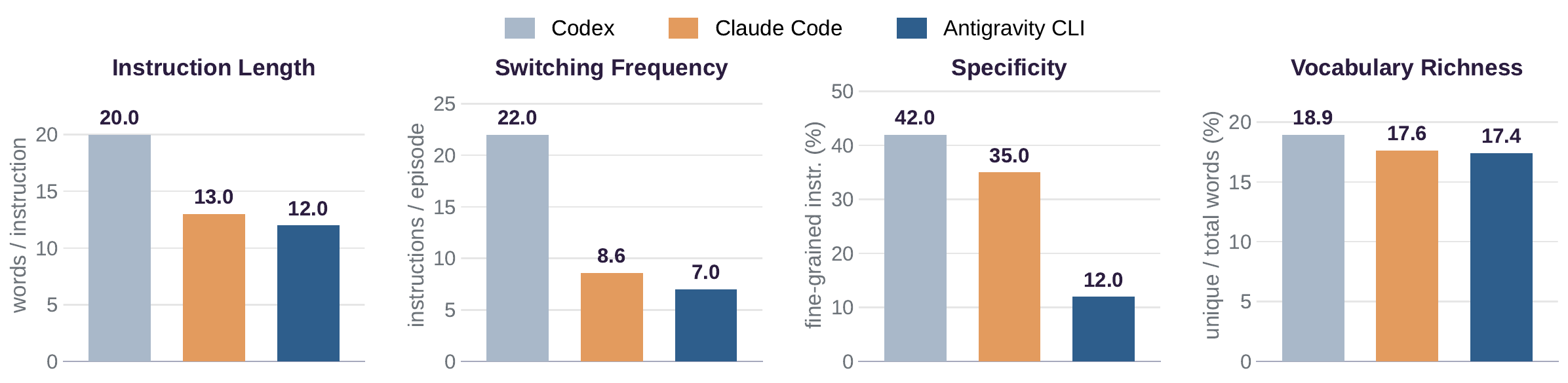}
    \caption{Planner instruction characteristics across three coding agents,
based on 10 randomly sampled open-ended episodes per agent.
Instruction length: average words per instruction;
switching frequency: instruction changes per episode;
specificity: proportion of instructions containing fine-grained execution details;
vocabulary richness: ratio of unique words to total words.}
    \label{fig:instruction_style}
\end{figure*}

\paragraph{Which Coding Agent Makes a Better Planner?} We select the three best-performing VLA models, Xiaomi Robotics-1,
InternVLA-A1.5, and $\pi_{0.5}$, along with the best-performing WAM,
OpenWAM-Alpha, as executors and pair each with different coding-agent
planners. As shown in Figure~\ref{fig:agent_performance}, Antigravity CLI
yields the largest performance gains across the evaluated policy executors.
To investigate these differences, we randomly sample 10 open-ended
episodes per planner and systematically compare their instruction styles
in Figure~\ref{fig:instruction_style}.
Antigravity CLI produces shorter instructions and switches them less
frequently. Its instructions also contain fewer execution-specific
details, such as grasping strategies or hand selection, and use simpler
vocabulary. In contrast, Codex often specifies detailed manipulation
procedures, including which hand to use.
These observations suggest that concise, high-level instructions may
be easier for VLA/WAM-based executors to follow, potentially because
they better match the language encountered during training.
Longer, more detailed instructions may instead introduce complexity
that these policies struggle to interpret.

\paragraph{Which Policy Model Makes a Better Executor?} As shown in Figure~\ref{fig:agent_performance}, InternVLA-A1.5 achieves
the best performance when paired with the same planner.
This advantage stems in part from its stronger manipulation capabilities,
especially robustness and failure recovery, which make it well suited
to serving as a policy executor.
We also observe that InternVLA-A1.5 follows revised planner instructions
more reliably during replanning and error recovery.
This behavior may be related to its extensive VQA training,
which helps preserve its language understanding capabilities.

\begin{figure*}[t]
    \centering
    \includegraphics[width=0.9\linewidth]{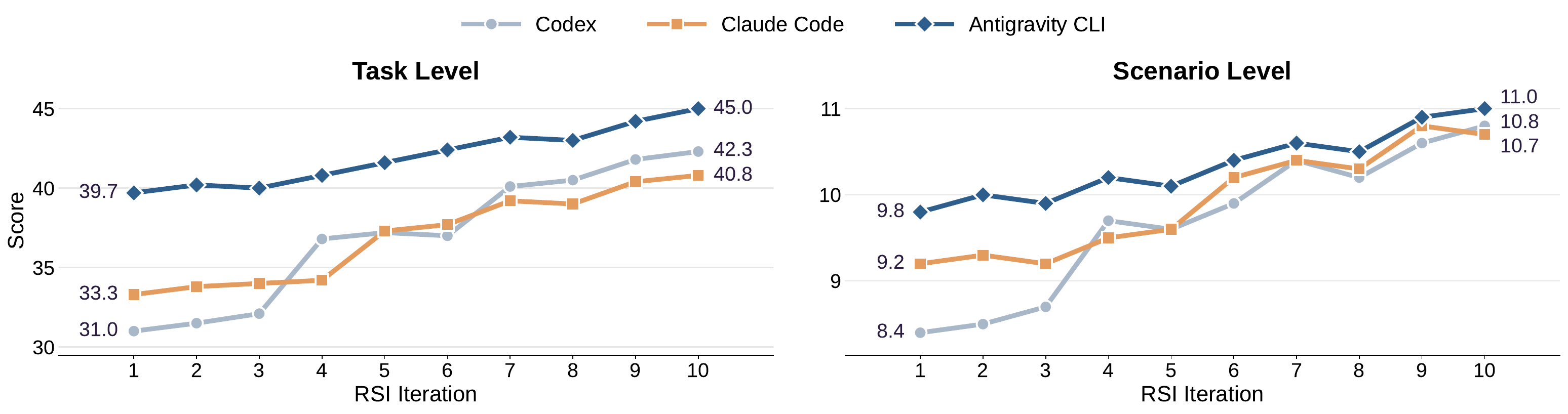}
    \caption{Task-level and scenario-level performance of three coding agents over successive RSI iterations}
    \label{fig:RSIResults}
\end{figure*}

\paragraph{Does RSI Improve Robot Agent Performance?} We evaluate three fully code-based agents, Codex, Claude Code, and Antigravity CLI, over ten RSI iterations, with 30 evaluation instances per task or scenario in each iteration.
Agents can accumulate experience across tasks and refine their tools,
implemented as reusable control functions.
As shown in Figure~\ref{fig:RSIResults}, all three agents improve
at both the task and scenario levels over successive RSI iterations,
while the gaps between Codex and the other two agents gradually narrow. In their initial attempts, agents often make basic errors, such as attempting to reach a target pose in a single motion that violates joint limits. Over successive iterations, they refine their control functions to decompose such motions into smaller steps through feasible intermediate poses.

\subsection{Sim--Real Performance Gap}

\begin{figure}[!htbp]
    \centering
    \includegraphics[width=\linewidth]{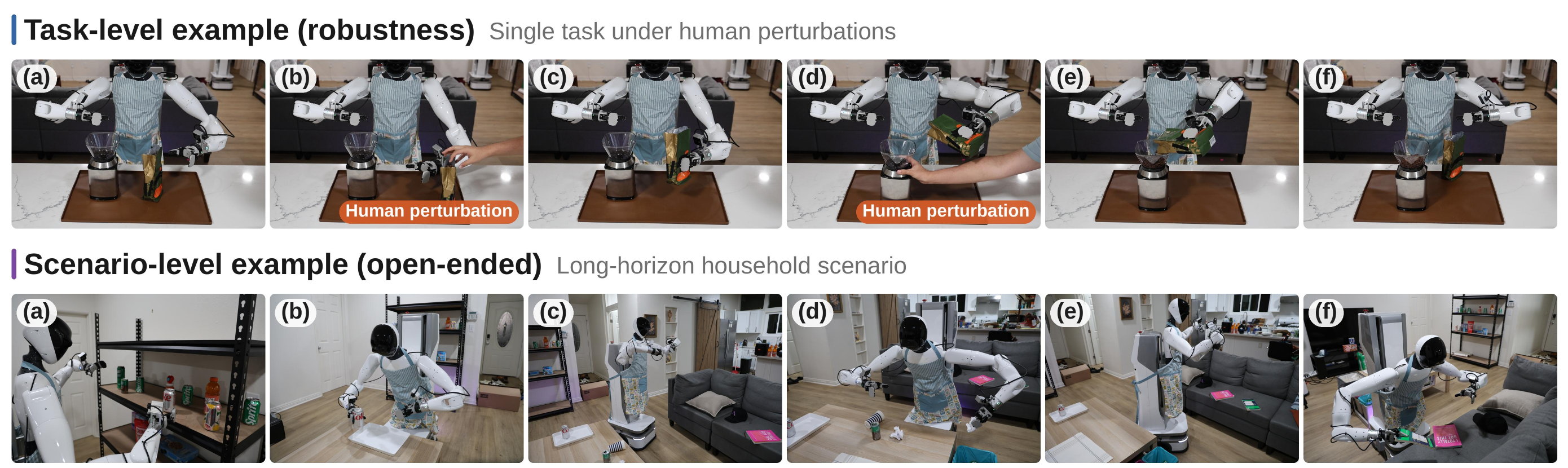}
    \caption{Real-world evaluation examples with the AgiBot G1.}
    \label{fig:real-openended-case-study}
\end{figure}

To assess performance differences between simulation and reality, we evaluate a subset of tasks and four representative scenario contexts, one from each open-ended scenario category, in both environments. We match the number of real-world and simulation training demonstrations for these tasks, using 50 demonstrations per atomic action and short-horizon task, and 10 per long-horizon task and open-ended scenario. We evaluate InternVLA-A1.5, OpenWAM-Alpha, and Codex powered by GPT-6 Astra. Each system is evaluated on 30 instances per task or scenario in each environment; the training-demonstration counts above are separate from these evaluation trials. As shown in Table~\ref{tab:sim_real}, both standalone models and the coding agent generally exhibit lower performance in the real world, reflecting the greater complexity of physical interactions and the increased visual clutter in real environments. The paired results characterize sim--real performance changes for the evaluated systems and matched task contexts. Figure~\ref{fig:real-openended-case-study} presents representative
case studies from the real-world evaluation.

\input{tables/table2}

%% file: tables/table1.tex
% Requires: booktabs, multirow, array, graphicx, tikz,
%           xcolor (with table option).
% Keep your existing ArenaSection color definition.

% Add these definitions to the preamble only once.
\definecolor{ArenaGold}{RGB}{218,174,55}
\definecolor{ArenaSilver}{RGB}{170,180,192}
\definecolor{ArenaBronze}{RGB}{190,133,86}

\newcommand{\ArenaMedal}[2]{%
    \tikz[baseline=(medal.base)]{
        \node[
            circle,
            fill=#1,
            inner sep=0pt,
            minimum size=1.2em,
            font=\tiny\bfseries,
            text=white
        ] (medal) {#2};
    }%
}
\newcommand{\ArenaGoldMedal}{\ArenaMedal{ArenaGold}{1}}
\newcommand{\ArenaSilverMedal}{\ArenaMedal{ArenaSilver}{2}}
\newcommand{\ArenaBronzeMedal}{\ArenaMedal{ArenaBronze}{3}}
\newcommand{\ArenaScore}[2]{%
    #1\,\makebox[1.4em][c]{#2}}

\begin{table*}[!h]
\centering
\caption{
Performance on RobotAgentBench. Discovery, Reasoning, and Recovery denote Task Discovery, Context Reasoning, and Failure Recovery, respectively. Average is the mean of the five task-level scores. Gold, silver, and bronze circles mark the top three scores in Average and Open-Ended, ranked separately for standalone models and robot agents. Agent architectures follow the P/E/V notation defined in
Section~\ref{sec:agent_arc}, where C denotes a coding agent,
L a learned VLA/WAM-based policy, and -- an absent component.
}
\label{tab:main_results}

\begingroup
\scriptsize
\setlength{\tabcolsep}{3pt}
\setlength{\extrarowheight}{0pt}
\renewcommand{\arraystretch}{1.08}
\def\ArenaRowStrut{\rule[-4pt]{0pt}{15pt}}
\setlength{\aboverulesep}{0pt}
\setlength{\belowrulesep}{0pt}

\resizebox{\textwidth}{!}{%
\begin{tabular}{
    @{}
    >{\ArenaRowStrut}l
    l
    ccccc
    >{\bfseries}c
    >{\bfseries}c
    @{}
}
\toprule
&
& \multicolumn{6}{c}{\ArenaRowStrut\textbf{Task-Level}}
& \textbf{Scenario-Level} \\
\cmidrule(lr){3-8}
\cmidrule(lr){9-9}
\textbf{Setting}
& \textbf{Model / Agent}
& \textbf{Manipulation}
& \textbf{Discovery}
& \textbf{Reasoning}
& \textbf{Recovery}
& \textbf{Robustness}
& \textbf{Average}
& \textbf{Open-Ended} \\
\midrule

\rowcolor{ArenaSection}[0pt][0pt]
\multicolumn{9}{@{}l@{}}{
    \ArenaRowStrut\textbf{Standalone VLA/WAM-based Policies}
} \\
\midrule

\multirow{8}{*}{VLA}
& $\pi_{0.5}$
& 13.1 & 1.2 & 16.7 & 20.3 & 8.1
& \ArenaScore{11.9}{\ArenaBronzeMedal}
& \ArenaScore{1.1}{} \\
& Xiaomi Robotics-1
& 18.1 & 0.5 & 15.8 & 5.8 & 21.6
& \ArenaScore{12.4}{\ArenaSilverMedal}
& \ArenaScore{0.4}{} \\
& InternVLA-A1.5
& 11.3 & 0.7 & 10.8 & 23.2 & 44.2
& \ArenaScore{18.0}{\ArenaGoldMedal}
& \ArenaScore{1.0}{} \\
& MolmoAct2
& 4.3 & 1.7 & 5.0 & 2.7 & 2.4
& \ArenaScore{3.2}{}
& \ArenaScore{1.9}{\ArenaSilverMedal} \\
& LingBot-VLA2
& 6.5 & 0.3 & 1.7 & 6.6 & 10.2
& \ArenaScore{5.1}{}
& \ArenaScore{0.5}{} \\
& HY-VLA
& 10.6 & 1.0 & 15.0 & 3.3 & 7.6
& \ArenaScore{7.5}{}
& \ArenaScore{0.8}{} \\
& Galaxea G0.5
& 11.0 & 1.5 & 25.0 & 9.3 & 8.1
& \ArenaScore{11.0}{}
& \ArenaScore{1.3}{\ArenaBronzeMedal} \\
& X-VLA
& 9.9 & 0.4 & 5.0 & 6.8 & 7.6
& \ArenaScore{5.9}{}
& \ArenaScore{0.3}{} \\
\midrule

\multirow{3}{*}{WAM}
& $\tau_0$ (low-level)
& 7.7 & 0.6 & 1.7 & 3.5 & 7.8
& \ArenaScore{4.3}{}
& \ArenaScore{0.5}{} \\
& OpenWAM-Alpha
& 13.4 & 1.2 & 15.0 & 4.6 & 8.6
& \ArenaScore{8.6}{}
& \ArenaScore{2.0}{\ArenaGoldMedal} \\
& LingBot-VA
& 5.0 & 1.5 & 3.3 & 5.4 & 4.2
& \ArenaScore{3.9}{}
& \ArenaScore{0.4}{} \\
\midrule

\rowcolor{ArenaSection}[0pt][0pt]
\multicolumn{9}{@{}l@{}}{
    \ArenaRowStrut\textbf{Robot Agents}
} \\
\midrule

\multirow{3}{*}{C/C/C}
& Codex
& 32.4 & 35.2 & 38.3 & 22.0 & 27.1
& \ArenaScore{31.0}{}
& \ArenaScore{8.4}{} \\
& Claude Code
& 20.6 & 24.5 & 60.0 & 44.2 & 17.1
& \ArenaScore{33.3}{}
& \ArenaScore{9.2}{} \\
& Antigravity CLI
& 29.2 & 36.5 & 80.0 & 27.8 & 25.0
& \ArenaScore{39.7}{\ArenaBronzeMedal}
& \ArenaScore{9.8}{\ArenaBronzeMedal} \\
\midrule

–/L/C
& Codex + InternVLA-A1.5
& 18.9 & 4.3 & 23.3 & 24.6 & 45.4
& \ArenaScore{23.3}{}
& \ArenaScore{3.7}{} \\

C/L/–
& Codex + InternVLA-A1.5
& 16.5 & 4.0 & 40.0 & 26.8 & 44.8
& \ArenaScore{26.4}{}
& \ArenaScore{9.1}{} \\

C/H/C
& Codex + InternVLA-A1.5
& 42.5 & 37.2 & 73.3 & 38.6 & 50.8
& \ArenaScore{48.5}{\ArenaGoldMedal}
& \ArenaScore{15.4}{\ArenaGoldMedal} \\

C/C/C + RSI
& Codex
& 36.8 & 33.3 & 63.3 & 31.4 & 46.5
& \ArenaScore{42.3}{\ArenaSilverMedal}
& \ArenaScore{10.8}{\ArenaSilverMedal} \\

\bottomrule
\end{tabular}%
}
\endgroup
\end{table*}

%% file: tables/table3.tex
\begingroup
\scriptsize
\setlength{\tabcolsep}{4pt}
\renewcommand{\arraystretch}{1.15}

\resizebox{\textwidth}{!}{%
\begin{tabular}{@{}llcccccccc@{}}
\toprule
&
& \multicolumn{4}{c}{\textbf{Scenario Progress}}
& \multicolumn{4}{c}{\textbf{Behavioral Metrics}} \\
\cmidrule(lr){3-6}
\cmidrule(lr){7-10}
\textbf{Setting}
& \textbf{Model / Agent}
& \textbf{Guests} $\uparrow$
& \textbf{Dinner} $\uparrow$
& \textbf{Leaving} $\uparrow$
& \textbf{Tidying} $\uparrow$
& \textbf{Coverage} $\uparrow$
& \textbf{Progress} $\uparrow$
& \textbf{Recovery} $\uparrow$
& \textbf{Stagnation} $\downarrow$ \\
\midrule

VLA
& InternVLA-A1.5
& 0.3 & 1.8 & 1.2 & 0.7
& 2.3 & 4.5 & 0.6 & 88.3 \\

WAM
& OpenWAM-Alpha
& 0.9 & 3.4 & 2.1 & 1.6
& 3.1 & 2.7 & 0.2 & 84.6 \\
\midrule

C/C/C
& Codex
& 11.2 & 3.6 & 1.4 & 17.4
& 13.4 & 15.2 & 15.4 & 33.7 \\

C/C/C
& Claude Code
& 6.4 & 12.8 & 8.1 & 9.5
& 12.8 & 16.2 & 18.8 & 18.7 \\

C/C/C
& Antigravity CLI
& 6.0 & 13.8 & 13.4 & 6.0
& 15.8 & 17.0 & 37.5 & 18.4 \\
\midrule

C/C/C + RSI
& Codex
& 13.1 & 6.4 & 3.0 & \textbf{20.7}
& \textbf{20.3} & 19.6 & \textbf{38.9} & 12.3 \\

C/H/C
& Codex + InternVLA-A1.5
& \textbf{15.8} & \textbf{14.3} & \textbf{13.9} & 17.6
& 15.2 & \textbf{25.1} & 30.7 & \textbf{12.1} \\

\bottomrule
\end{tabular}%
}
\endgroup

%% file: tables/table2.tex
% Preamble: add only once.
% Requires: booktabs, multirow, array, graphicx, xcolor
\definecolor{SimRealUp}{RGB}{34,139,75}
\definecolor{SimRealDown}{RGB}{190,55,55}

% Arrows occupy no horizontal space.
\newcommand{\RealUp}[1]{%
    #1\makebox[0pt][l]{\,{\color{SimRealUp}$\uparrow$}}}
\newcommand{\RealDown}[1]{%
    #1\makebox[0pt][l]{\,{\color{SimRealDown}$\downarrow$}}}

\begin{table*}[!h]
\centering
\caption{
Performance comparison between simulation and the real world
on matched evaluation subsets.
Green upward and red downward arrows indicate increases and
decreases relative to simulation, respectively.
}
\label{tab:sim_real}

\begingroup
\scriptsize
\setlength{\tabcolsep}{6pt}
\renewcommand{\arraystretch}{1.15}

\resizebox{\textwidth}{!}{%
\begin{tabular}{
    @{}ll
    rrrrr
    >{\bfseries}r
    >{\bfseries}r
    @{\hspace{0.8em}}
}
\toprule
& & \multicolumn{6}{c}{\textbf{Task-Level}}
& \multicolumn{1}{c}{\textbf{Scenario-Level}} \\
\cmidrule(lr){3-8}
\cmidrule(lr){9-9}
\textbf{Model / Agent}
& \textbf{Environment}
& \textbf{Manipulation}
& \textbf{Discovery}
& \textbf{Reasoning}
& \textbf{Recovery}
& \textbf{Robustness}
& \textbf{Average}
& \textbf{Open-Ended} \\
\midrule

\multirow{2}{*}{InternVLA-A1.5}
& Sim
& 11.8
& 1.2
& 12.5
& 20.8
& 42.7
& 17.8
& 1.4 \\
& Real
& \RealDown{10.7}
& \RealUp{1.7}
& \RealDown{10.0}
& \RealDown{15.0}
& \RealDown{30.0}
& \RealDown{13.5}
& \RealDown{1.0} \\
\midrule

\multirow{2}{*}{OpenWAM-Alpha}
& Sim
& 14.5
& 0.8
& 12.5
& 5.5
& 9.2
& 8.5
& 2.4 \\
& Real
& \RealDown{10.0}
& \RealDown{0.0}
& \RealUp{15.0}
& \RealDown{5.3}
& \RealUp{10.5}
& \RealDown{8.2}
& \RealDown{0.0} \\
\midrule

\multirow{2}{*}{Codex}
& Sim
& 35.8
& 38.4
& 45.8
& 24.1
& 26.8
& 34.2
& 8.7 \\
& Real
& \RealDown{33.0}
& \RealDown{33.3}
& \RealDown{44.2}
& \RealUp{25.0}
& \RealDown{24.0}
& \RealDown{31.9}
& \RealDown{6.0} \\

\bottomrule
\end{tabular}%
}
\endgroup
\end{table*}

%% file: sections/related_works.tex
\section{Related Works}
\label{related}

\paragraph{Robot Manipulation Benchmarks.}
Robot manipulation benchmarks evaluate increasingly diverse and complex tasks, yet largely retain predefined goals and task-specific demonstrations. RLBench~\citep{rlbench}, Meta-World~\citep{metaworld}, ManiSkill2~\citep{maniskill2}, and LIBERO~\citep{libero} assess individual manipulation skills through dedicated tasks, while RoboCasa~\citep{robocasa}, BEHAVIOR~\citep{srivastava2022behavior, li2023behavior}, and RoboTwin~\citep{mu2024robotwin, robotwin2} extend coverage to diverse household activities and environments. VLABench~\citep{vlabench} and LIBERO-Long~\citep{libero} extend evaluation to longer-horizon tasks, providing complete demonstrations that allow policies to learn entire task sequences through imitation. More recent benchmarks, including RoboDojo~\citep{chen2026robodojo}, RoboCerebra~\citep{rocerebra}, and RoboMME~\citep{dai2026robomme}, assess reasoning, planning, and memory alongside physical execution, but their narrowly specified tasks offer limited coverage of context-dependent household assistance. \textit{RobotAgentBench} evaluates diverse agentic capabilities through task-level assessments and open-ended household scenarios, with training demonstrations primarily teaching reusable skills that agents must select and compose according to the current situation.

\paragraph{Embodied Planning Benchmarks.}
Beyond robot manipulation benchmarks, some work abstracts physical execution into predefined callable functions, focusing on evaluating the planning and reasoning capabilities of models in simulated embodied activities.
ALFRED~\citep{alfred}, ALFWorld~\citep{alfworld}, RealFRED~\citep{realfred}, and EmbodiedBench~\citep{embodiedbench} assess household instruction following and planning, while WorldLines~\citep{worldlines} emphasizes persistent world state and memory. TEACh~\citep{teach}, DialFRED~\citep{dialfred}, KnowNo~\citep{knowno}, AmbiK~\citep{ambik}, and Ask-to-Act~\citep{asktoact} further evaluate interaction and clarification under incomplete or ambiguous instructions. However, by abstracting away physical control, these benchmarks provide a limited assessment of how planning and reasoning must adapt to execution outcomes, particularly when recovering from failures in real-world tasks.

\paragraph{Robot Agents} Robot agents integrate reasoning and physical execution through different architectures. SayCan~\citep{ahn2022can}, Inner Monologue~\citep{huang2022inner}, and Hi Robot~\citep{shi2025hi} combine high-level planning by language or vision-language models with learned skills or policies. Code as Policies~\citep{liang2023code}, ProgPrompt~\citep{singh2022progprompt}, Instruct2Act~\citep{huang2023instruct2act}, and VoxPoser~\citep{huang2023voxposer} use generated programs to connect perception, planning, and control. Harness VLA~\citep{zhang2026harness} combines learned policies with analytic control primitives within an agentic framework. Recent approaches further enable execution-time correction of policy actions~\citep{team2026systematically} and feedback-driven improvement of tools and control code~\citep{physicalrsi2026, ge2026make}. \textit{RobotAgentBench} provides a unified benchmark for systematically evaluating these robot agent designs.

%% file: appendix/appendix.tex
% Benchmark appendix. All task entries use the same RAATask macro.
\setlength{\parskip}{2pt}
\setlength{\parindent}{0pt}
\raggedbottom
\captionsetup{font=footnotesize,labelfont=bf,skip=3pt,hypcap=false}
\newcommand{\RAATask}[9]{%
  \par\addvspace{4pt}\noindent\begin{minipage}{\linewidth}
  \setlength{\parskip}{2pt}%
  \subsection[#1 #2]{#1\quad #2}
  \textbf{Instruction.} #3\par
  \textbf{Initial state.} #4\par\vspace{2pt}
  \noindent\includegraphics[width=\linewidth]{#8}\par
  \captionof{figure}{#1. #9}\label{fig:raa-#1}\vspace{2pt}
  \textbf{Procedure.} #5\par
  \textbf{Completion.} #6\par
  \textbf{Focus.} #7\par
  \end{minipage}\par\addvspace{4pt}%
}
\section{Benchmark Setup}
\label{app:benchmark_setup}

This section describes the information available to evaluated systems, the robot action space, and the shared simulation settings. Task and scenario definitions are provided in Appendix~\ref{app:task_details}.

\subsection{Provided Information}

The evaluation interface provides all systems with access to the same information sources: task instructions and context, camera observations, robot-state feedback, and a progressively revealed navigation map. The same information disclosure rules apply to all systems. How this information is represented, integrated, and used depends on the model architecture, input adapter, and agent design. Model-specific inputs are described in Appendix C.

\paragraph{Task instructions and context.}
Each episode provides a task instruction or a high-level scenario request. Context-dependent settings additionally supply information such as user preferences, household routines, and storage arrangements. The information provided for each task family is described in Appendix~\ref{app:task_details}.

\paragraph{Visual observations.}
The benchmark records synchronized RGB observations from one head-mounted camera at $1280 \times 800$ and two wrist-mounted cameras at $848 \times 480$. Learned policies receive updated observations after executing an action prefix; the prediction horizon and executed prefix length are separate quantities. Code-based agents can request observations through the evaluation interface.

\paragraph{Robot state.}
We use the AgiBot G1 as the robot embodiment. Its state is represented by a 23-dimensional vector containing twenty joint channels and the mobile base's world-frame position $(x,y)$ and unwrapped yaw. The joint channels comprise fourteen arm joints, two torso/waist joints, two head joints, and two gripper channels. Torso lift is expressed in metres, while angular channels are expressed in radians. Table~\ref{tab:state_action_spaces} summarizes the state and action representations.

\paragraph{Navigation map.}
Map-assisted evaluations provide a read-only top-down navigation map at a resolution of 0.04\,m. The map is derived from static scene geometry rather than reconstructed from RGB images. It is progressively revealed using the measured head-camera pose, field of view, and static occlusion, with a small local region around the robot treated as known. The same generation and revelation rules apply across participating systems; revealed regions may differ because systems choose different viewpoints and trajectories.

The map distinguishes known free floor, static obstacles, and unexplored regions, and displays the robot's measured position, heading, and executed trajectory. The complete map, generated movable objects, task labels, and target poses are not exposed. No planned routes or automatic navigation actions are provided. Systems must choose and execute their own movements and use camera observations to assess movable objects.

\begin{figure}[t]
\centering
\includegraphics[width=\linewidth]{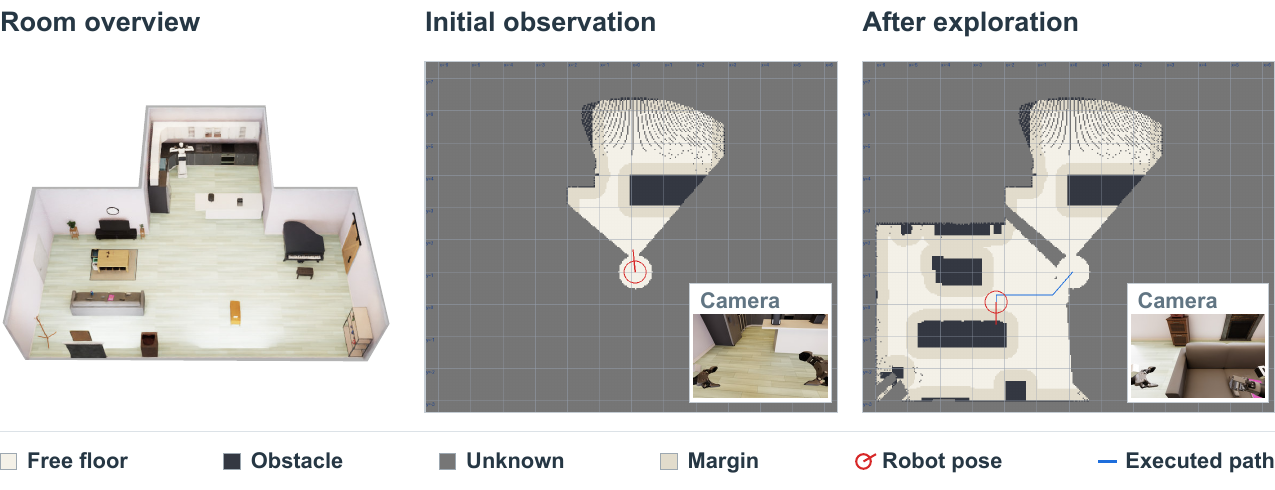}
\caption{Room overview (left) and progressively revealed maps with corresponding camera views (right). The room overview is for illustration and is not an agent input.}
\label{fig:map-observations}
\end{figure}

\subsection{Action Space}

Each action is a 22-dimensional whole-body command containing fourteen arm-joint targets, two torso/waist targets, two head-joint targets, two continuous gripper targets, and two mobile-base commands specifying forward velocity and yaw rate. Joint commands specify target positions, while base commands specify velocities. The channel definitions, units, and nominal ranges are summarized in Table~\ref{tab:state_action_spaces}.

\input{tables/table_appendix1}

\subsection{Simulation and Execution Settings}

\paragraph{Scene construction.}
We first construct a digital twin of a real home, matching its layout, furniture dimensions, and appearance. We then create additional layouts by modifying floor plans, including T-shaped and L-shaped configurations, and rearranging functional areas. All layouts are manually reviewed for geometric validity and plausible furniture arrangements. The resulting catalog contains 50 distinct layouts, with additional visual variation in furniture colours, textures, materials, and interior styles.

\paragraph{Physics and control.}
Robot parameters, camera calibration, asset properties, and scene settings are resolved before each episode and remain fixed across its subtasks. Physics integration runs at 60 or 240\,Hz, depending on the configuration, while the action interface operates at 20\,Hz. The integration rate and backend remain unchanged within an episode. Cloth manipulation may use a dedicated soft-body solver. Grasping relies on physical contact and friction between the gripper pads and objects.

\paragraph{Continuous execution and scoring.}
Task transitions do not reset the scene, object states, or accumulated progress. Task processes and appliance dynamics persist as the robot moves between workspaces. Progress is tracked throughout the episode, while placement credit is updated using the current physical state, so disturbing an earlier placement can reduce its final credit. At episode termination, all registered tasks are finalized using their task-specific scoring rules. Task-specific completion criteria are provided in Appendix~\ref{app:task_details}.

% Task Details grouped by evaluation capability.
\input{appendix/task_details}

\section{Policy training and model-specific adaptation}
\label{app:policy_training}

This section describes how we fine-tune and adapt the eleven
policy baselines to the AgiBot G1, covering shared training
settings and model-specific configurations.

\subsection{Shared training setup}

\paragraph{Training data and split.} We fine-tune all eleven policy baselines using demonstrations from the 101.8-hour RobotAgentBench simulation dataset, covering atomic actions, short tasks, long tasks, and open-ended household scenarios. All models use the same episode-level training/validation split.

\paragraph{Data representation and preprocessing.}
Demonstrations follow a unified format with 23-dimensional robot
states and 22-dimensional whole-body actions, as defined in
Table~\ref{tab:state_action_spaces}.
We adapt the data to each model's required input format,
including image resizing, normalization, and multi-view arrangement.
State and action normalization statistics are computed
from the training split.
Image augmentations are applied only during training.
Model-specific adaptations are described below.

\paragraph{Fine-tuning and sampling.} We follow each model’s native fine-tuning recipe, retaining its prescribed frozen components. In selected continuation phases, 80\% of training samples come from a combined pool of short- and long-task demonstrations, and 20\% from atomic-action demonstrations.

\paragraph{Training-budget convention.} Table~\ref{tab:raa-training} reports trainable parameter counts, final-phase global batch sizes, and total optimizer updates. Global batch sizes include gradient accumulation, and each training step corresponds to one optimizer update.

\par\medskip\noindent\begin{minipage}{\linewidth}\footnotesize\setlength{\tabcolsep}{6pt}\renewcommand{\arraystretch}{1.15}\begin{tabularx}{\linewidth}{>{\raggedright\arraybackslash}Xrrr}\toprule Model & Trainable (B) & Final-phase batch & Optimizer updates \\ \midrule
$\pi_{0.5}$ & 3.35 & 1,024 & 16,000 \\
Xiaomi Robotics-1 & 4.72 & 56 & 70,000 \\
InternVLA-A1.5 & 2.68 & 64 & 120,000 \\
MolmoAct2 & 5.05 & 32 & 100,000 \\
LingBot-VLA2 & 6.38 & 256 & 60,000 \\
LingBot-VA & 5.09 & 128 & 16,000 \\
HY-VLA & 4.53 & 128 & 18,000 \\
Galaxea G0.5 & 2.86 & 128 & 28,000 \\
X-VLA & 0.88 & 768 & 33,500 \\
$\tau_0$ low-level policy & 2.50 & 256 & 35,000 \\
OpenWAM-Alpha & 6.02 & 1,280 & 49,203 \\
\bottomrule\end{tabularx}\captionof{table}{Fine-tuning configurations. Parameter counts cover the trainable policy scope in billions. Batch sizes describe the final phase, rather than the full training history.}\label{tab:raa-training}\end{minipage}\par\medskip

\subsection{Model-specific settings}

\paragraph{\texorpdfstring{$\pi_{0.5}$}{pi0.5}.}
We fine-tune from \path{pi05_base} for 16,000 optimizer updates,
with a final-phase global batch size of 1,024.
The G1 adapter maps the benchmark's state and action channels
to the model's native format.

\paragraph{Xiaomi Robotics-1.}
We fine-tune from the released Xiaomi Robotics-1 base checkpoint
for 70,000 optimizer updates with a global batch size of 56,
adapting the model to the G1 whole-body state and action interface.

\paragraph{InternVLA-A1.5.}
We fine-tune from \path{InternRobotics/InternVLA-A1.5-base}
for 120,000 optimizer updates with a global batch size of 64.
We retain the model's native training objectives and frozen Wan teacher.

\paragraph{MolmoAct2.}
We fine-tune from \path{allenai/MolmoAct2}
for 100,000 optimizer updates,
with a final-phase global batch size of 32.
Training retains both continuous and discrete action objectives,
with LoRA disabled.

\paragraph{LingBot-VLA2.}
We fine-tune from \path{robbyant/lingbot-vla-v2-6b}
for 60,000 optimizer updates,
with a final-phase global batch size of 256.
The adapter maps the benchmark's state and action channels
to the model's native representation.

\paragraph{LingBot-VA.}
\label{app:lingbot_va_training}
We fine-tune from \path{robbyant/lingbot-va-base}
for 16,000 optimizer updates with a global batch size of 128.
The policy conditions on RGB observations and language
without proprioceptive input, and its action output is adapted
to the 22 G1 command channels.

\paragraph{HY-VLA.}
We fine-tune from \path{tencent/Hy-Embodied-0.5-VLA-UMI}
for 18,000 optimizer updates,
with a final-phase global batch size of 128.
We retain its six-frame image-history mechanism,
sampling historical images every five source frames.

\paragraph{Galaxea G0.5.}
We fine-tune from the released G0.5 base policy
for 28,000 optimizer updates,
with a final-phase global batch size of 128.
We retain its released action-tokenizer codec and adapt
the whole-body channels to its native action representation.

\paragraph{X-VLA.}
We fine-tune from \path{2toINF/X-VLA-Pt}
for 33,500 optimizer updates,
with a final-phase global batch size of 768.
The backbone remains frozen for the first 1,000 updates,
following the initial training recipe.

\paragraph{\texorpdfstring{$\tau_0$ low-level policy}{tau0 low-level policy}.}
We fine-tune the low-level action policy from
\path{sii-research/tau-0-vla}
for 35,000 optimizer updates,
with a final-phase global batch size of 256.
This configuration evaluates the action policy
rather than the complete world-model hierarchy.

\paragraph{OpenWAM-Alpha.}
We fine-tune from the released OpenWAM-Alpha Pretrain Foundation Model
for 49,203 optimizer updates,
with a final-phase global batch size of 1,280.
Training retains the joint video-and-action objective
and frozen auxiliary encoders.
The three camera views are combined into a single image canvas.
\section{Detailed Analysis of Open-ended Scenarios}
\label{app:evaluation_reporting}

We complement scenario scores with behavioural metrics and an analysis of how agents distribute their interactions and respond to execution difficulties.

\subsection{Behavioural metrics}

\paragraph{Task Coverage.}
Task Coverage measures the proportion of required tasks physically engaged by the agent:
\[
\mathrm{Coverage}
=
100 \times
\frac{\sum_e |I_e|}{\sum_e |T_e|},
\]
where $T_e$ is the required task set in episode $e$ and $I_e \subseteq T_e$ is the engaged subset. Engagement requires eligible gripper-finger contact with a registered target or a task-specific operation-start event. Each task counts once; workspace selection and background monitoring do not count.

\paragraph{Task Progress.}
Task Progress measures process-score attainment among engaged tasks:
\[
\mathrm{Progress}
=
100 \times
\frac{\sum_e \sum_{t \in I_e} p_{e,t}}
     {\sum_e \sum_{t \in I_e} p^{\max}_{e,t}},
\]
where $p_{e,t}$ and $p^{\max}_{e,t}$ are the process score at termination and its maximum. Target-contact credit is included, while final-state score channels and unengaged tasks are excluded.

\paragraph{Recovery Rate.}
Recovery Rate is the percentage of observable physical-failure segments followed by task-specific success of the same task at a later control step. Repeated failures before recovery form one segment. API errors, formatting errors, and termination-only settlement events are excluded.

For these three metrics, numerators and denominators are pooled across scenarios before computing percentages. Zero denominators are reported as undefined.

\paragraph{Stagnation.}
Stagnation measures the fraction of simulation time spent in repeated physical execution without progress. An interval requires at least three executions that advance simulation; only time beyond a 30-second grace period counts. Task-score increases, manipulation progress, or newly explored map area interrupt the interval. Successful navigation exceeding 2\,cm translation or 0.035\,rad rotation is excluded.

State loops also qualify when a cycle of at most eight calls repeats at least three times over at least 30 seconds without task or map progress; the first cycle is excluded. Matching tolerances are 2\,cm for base position, 0.035\,rad for joints and base yaw, and 1\,cm/0.052\,rad for hand position/orientation. Overlapping intervals count once. Time follows the 20\,Hz simulation clock, excluding API waiting.

\subsection{Behavioural patterns and execution issues}

\begin{figure}[t]
\centering
\includegraphics[width=\linewidth]{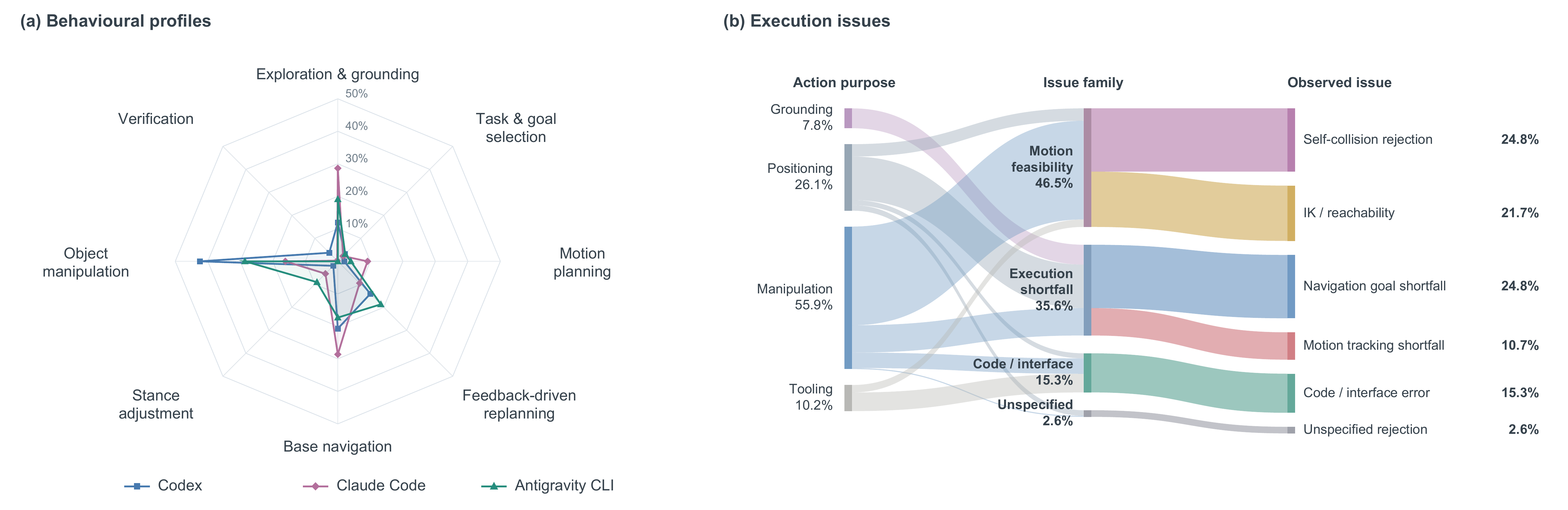}
\caption{Behavioural patterns and execution issues in open-ended scenarios. (a) Scenario-balanced task-event distributions. (b) Identified execution issues grouped by action purpose and issue family. Percentages are rounded independently.}
\label{fig:openend-error-analysis}
\end{figure}

\paragraph{Analysis protocol.}
We analyse interaction traces of Codex, Claude Code, and Antigravity CLI across the four open-ended scenarios. Figure~\ref{fig:openend-error-analysis}(a) groups submitted action attempts and deduplicated public planning statements into eight behavioural categories. Compound scripts count once, and verification includes only standalone checks. Task-event shares exclude setup and unclassified events and are averaged equally across scenarios after within-episode normalization.

Panel (b) summarizes identified execution issues. Calls are weighted inversely by each episode's submitted-action count, with equal agent and scenario weights before normalization over issue-bearing calls. Each call receives at most one issue label. Navigation shortfalls require residual translation exceeding 2\,cm or rotation exceeding 0.02\,rad. These percentages describe issue composition rather than the overall error rate.

\paragraph{Behavioural patterns.}
Object manipulation accounts for 42.4\% of Codex's task events, compared with 16.1\% for Claude Code and 28.6\% for Antigravity CLI. Claude Code instead concentrates more interactions on exploration, grounding, and base navigation, which together account for 57.3\% of its events.

For all three agents, feedback-driven replanning is more frequent than explicit task and goal selection. Observed adjustments include smaller approach steps and revised wrist orientations, illustrating the recurring need to update actions using execution feedback.

\paragraph{Execution difficulties.}
Motion-feasibility issues and execution shortfalls together account for 82.1\% of identified issues, compared with 15.3\% for code or interface errors. These difficulties span both positioning and object manipulation, suggesting that reliable execution requires more than correct tool calls.

Agents need to coordinate navigation, stance adjustment, and manipulation, and revise unsuccessful approaches using execution feedback. Such revisions should be followed by verification that the intended physical progress has occurred.

%% file: tables/table_appendix1.tex
\begin{table*}[!h]
\centering
\caption{
Robot state and action spaces. Joint entries represent measured
positions in the state and position targets in the action.
Base position and unwrapped yaw are expressed in the world frame;
base actions specify forward velocity and yaw rate.
Ranges are nominal command bounds, not empirical state bounds; the arm
intervals summarize envelopes over joint-specific limits. Base-speed
bounds describe the demonstration-collection controller.
}
\label{tab:state_action_spaces}

\begingroup
\small
\setlength{\tabcolsep}{5pt}
\renewcommand{\arraystretch}{1.12}

% 修改 0.90 即可调整整个表的大小
\resizebox{0.90\textwidth}{!}{%
\begin{tabular}{
    @{} clcc
    @{\hspace{24pt}}
    clcc @{}
}
\toprule
\multicolumn{4}{c}{\textbf{State space}\enspace (23 dimensions)}
&
\multicolumn{4}{c}{\textbf{Action space}\enspace (22 dimensions)}
\\
\midrule

Dim. & Component & Unit & Nominal range
& Dim. & Component & Unit & Nominal range \\

\midrule

0--6 & Left arm joints & rad & $[-\pi,\pi]$
& 0--6 & Left arm joints & rad & $[-\pi,\pi]$ \\

7--13 & Right arm joints & rad & $[-\pi,\pi]$
& 7--13 & Right arm joints & rad & $[-\pi,\pi]$ \\

14 & Torso lift & m & $[0,0.55]$
& 14 & Torso lift & m & $[0,0.55]$ \\

15 & Waist pitch & rad & $[0,\pi/2]$
& 15 & Waist pitch & rad & $[0,\pi/2]$ \\

16 & Head yaw & rad & $[-\pi/2,\pi/2]$
& 16 & Head yaw & rad & $[-\pi/2,\pi/2]$ \\

17 & Head pitch & rad & $[-0.349,0.524]$
& 17 & Head pitch & rad & $[-0.349,0.524]$ \\

18 & Left gripper & rad & $[0,\pi/4]$
& 18 & Left gripper & rad & $[0,\pi/4]$ \\

19 & Right gripper & rad & $[0,\pi/4]$
& 19 & Right gripper & rad & $[0,\pi/4]$ \\

\midrule

20 & Base $x$ & m & ---
& 20 & Base forward velocity & m/s & $[-0.18,0.18]$ \\

21 & Base $y$ & m & ---
& 21 & Base yaw rate & rad/s & $[-0.25,0.25]$ \\

22 & Base yaw (unwrapped) & rad & ---
& & & & \\

\bottomrule
\end{tabular}%
}

\endgroup
\end{table*}

%% file: appendix/task_details.tex
% Replacement confined to the existing Task Details subsection.
\section{Task and Scenario Descriptions}\label{app:task_details}
\label{app:task-details}
\newenvironment{RAADetail}[1]{%
  \par\addvspace{5pt}\noindent\begin{minipage}{\linewidth}%
  \setlength{\parskip}{2pt}\setlength{\parindent}{0pt}%
  \captionsetup{font=footnotesize,labelfont=bf,skip=2pt,hypcap=false,justification=raggedright,singlelinecheck=false}%
  {\bfseries #1}\par\nobreak
}{\end{minipage}\par\addvspace{5pt}}
\newenvironment{RAAFlow}[1]{%
  \par\addvspace{5pt}\begingroup\setlength{\parskip}{2pt}%
  \clubpenalty=10000\widowpenalty=10000%
  \captionsetup{font=footnotesize,labelfont=bf,skip=2pt,hypcap=false,justification=raggedright,singlelinecheck=false}%
  \noindent\textbf{#1}\par\nobreak
}{\par\endgroup\addvspace{5pt}}

The following sections describe the tasks and scenarios used to evaluate robot autonomy. Input descriptions refer to the environment interface; individual policy inputs follow the adapters in Appendix~\ref{app:policy_training}. Manipulation includes 18 short tasks and 8 long tasks. Task discovery and context reasoning assess whether agents identify relevant objects through gripper contact. Failure recovery evaluates execution from predefined failure states, while robustness evaluates adaptation to disturbances during execution. Four open-ended household scenarios combine task discovery, reasoning, and manipulation under high-level user requests. The figures illustrate task settings and procedures using demonstrations and diagnostic renders, rather than evaluated-agent results.

\subsection{Manipulation}
\textbf{Task overview.} We evaluate manipulation through 18 short tasks and 8 long tasks, covering both individual manipulation skills and their composition into more complex household activities.\par

\textbf{Input and setting.} The agent receives a task instruction, camera observations, and robot-state feedback. Each episode starts from a task-specific scene. Long tasks require continuous execution across subgoals without resetting the environment.\par
\textbf{Completion and scoring.} Performance is measured using task-specific process scores. All 18 short tasks and 8 long tasks contribute to the manipulation score.\par

\par\medskip\noindent\textbf{Short tasks}\par\nopagebreak[4]
\begin{RAADetail}{1. Prepare coffee}
\textbf{Request.} Prepare a cup of coffee.\par
\textbf{Setting.} A coffee machine, an empty cup, and a bag of coffee beans are on the countertop.\par
\textbf{Task.} Place the cup upright under the coffee outlet and release it. Pour beans into the hopper, then return the bag upright to the countertop and release it. Press the button to start brewing, press it again to stop dispensing, and retrieve the same cup.\par
\textbf{Completion.} Prepare the coffee and retrieve the cup while keeping it upright and avoiding excessive spillage.\par
\par\noindent\begin{minipage}{\linewidth}
\includegraphics[width=\linewidth]{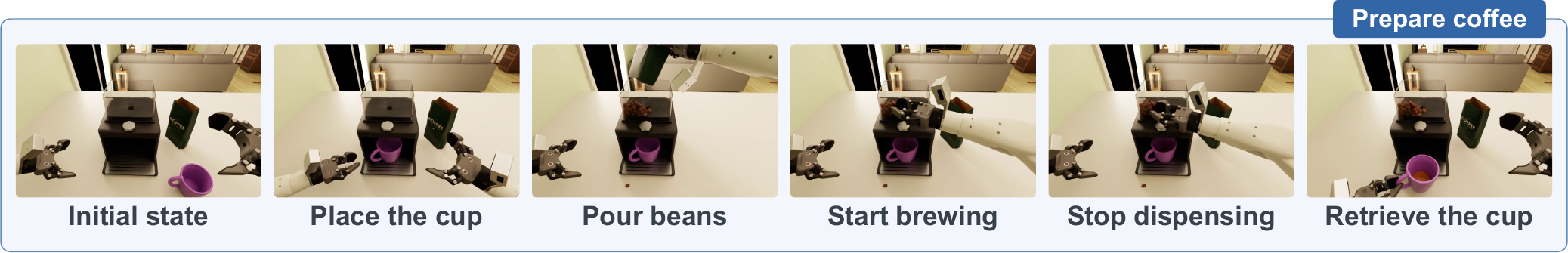}
\captionof{figure}{Procedure for the ``Prepare coffee'' task.}\label{fig:task-01}
\end{minipage}\par
\end{RAADetail}

\begin{RAADetail}{2. Load and unload the air fryer}
\textbf{Request.} Load the designated pastry into the air fryer, then take it out.\par
\textbf{Setting.} The air fryer basket is initially closed. Two pastries are on a plate, one of which is designated as the target.\par
\textbf{Task.} Pull out the basket, place the target pastry inside, and push the basket closed. Then remove the same basket completely, set it down securely, and retrieve the same pastry. Heating is not required.\par
\textbf{Completion.} Load the target pastry, remove and set down the basket, and retrieve the pastry.\par
\par\noindent\begin{minipage}{\linewidth}
\includegraphics[width=\linewidth]{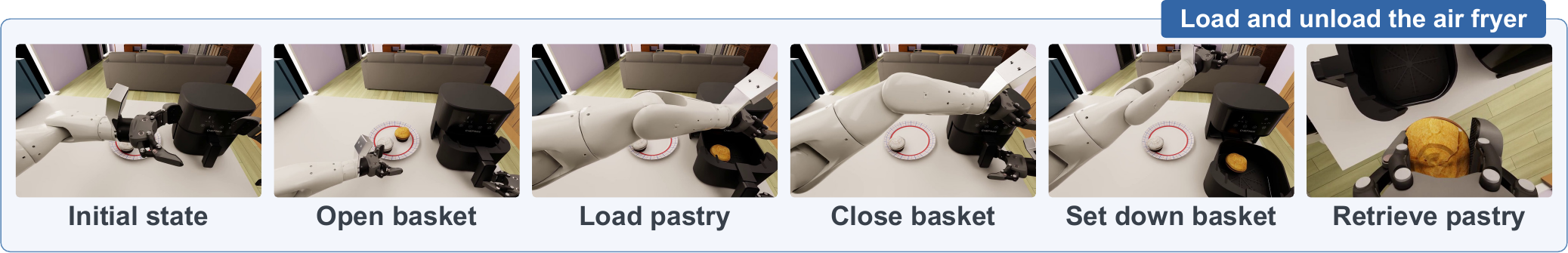}
\captionof{figure}{Procedure for the ``Load and unload the air fryer'' task.}\label{fig:task-02}
\end{minipage}\par
\end{RAADetail}

\begin{RAADetail}{3. Stock the tray}
\textbf{Request.} Bring the requested product from the shelf to the small wooden tray.\par
\textbf{Setting.} The target product is on the snack shelf, and the small wooden tray is on the coffee table.\par
\textbf{Task.} Pick up the requested product, carry it to the coffee table, and place it in the small wooden tray.\par
\textbf{Completion.} Leave the requested product in the small wooden tray after releasing it.\par
\par\noindent\begin{minipage}{\linewidth}
\includegraphics[width=\linewidth]{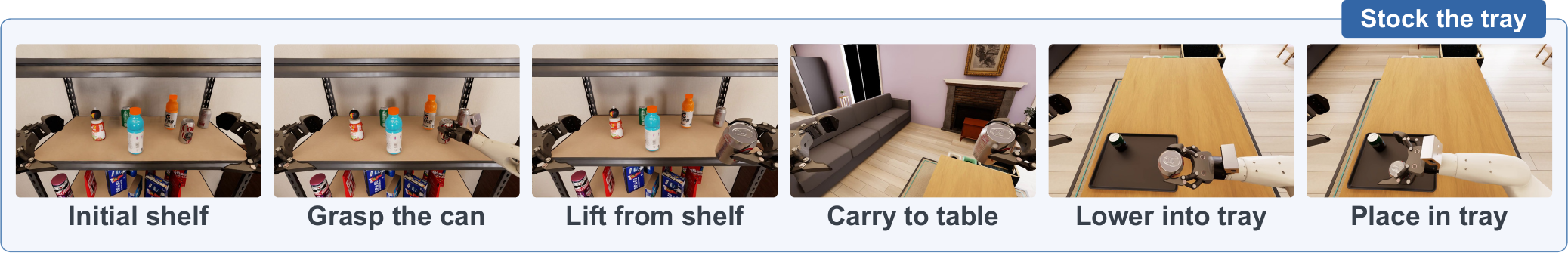}
\captionof{figure}{Procedure for the ``Stock the tray'' task.}\label{fig:task-03}
\end{minipage}\par
\end{RAADetail}

\begin{RAADetail}{4. Retrieve items from the drawer}
\textbf{Request.} Take the requested object out of the drawer and put it in the specified container.\par
\textbf{Setting.} A wallet or sunglasses case is inside the initially closed coffee-table drawer. The destination is specified as either the open tote bag or the item-storage basket.\par
\textbf{Task.} Open the drawer, grasp the target, lift it fully clear of the drawer rim, carry it to the specified container, and release it inside. Closing the drawer afterward is not required.\par
\textbf{Completion.} Complete the retrieval and leave the same object released inside and supported by the specified container.\par
\par\noindent\begin{minipage}{\linewidth}
\includegraphics[width=\linewidth]{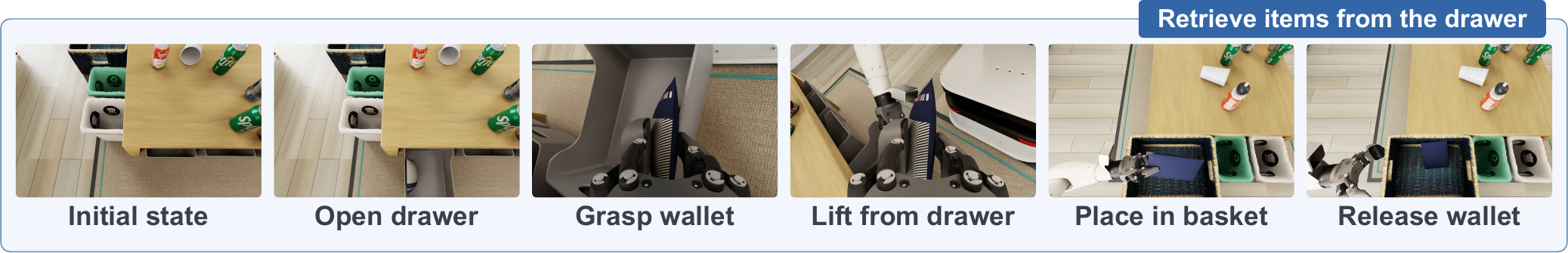}
\captionof{figure}{Procedure for the ``Retrieve items from the drawer'' task.}\label{fig:task-04}
\end{minipage}\par
\end{RAADetail}

\begin{RAADetail}{5. Store items in the drawer}
\textbf{Request.} Put the requested sofa object into the coffee-table drawer and close the drawer.\par
\textbf{Setting.} A wallet or green notebook is on the sofa, and the coffee-table drawer is initially closed.\par
\textbf{Task.} Pick up the requested object from the sofa, carry it to the coffee table, open the drawer, place and release the object inside, and close the drawer.\par
\textbf{Completion.} The drawer must be closed, with the same object released inside and supported by its interior.\par
\par\noindent\begin{minipage}{\linewidth}
\includegraphics[width=\linewidth]{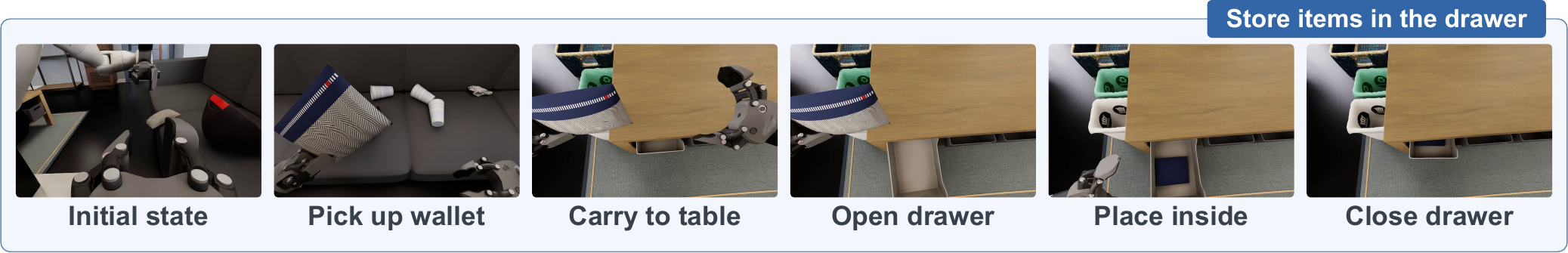}
\captionof{figure}{Procedure for the ``Store items in the drawer'' task.}\label{fig:task-05}
\end{minipage}\par
\end{RAADetail}

\begin{RAADetail}{6. Hang the cap}
\textbf{Request.} Take the cap from the sofa and hang it on the coat rack.\par
\textbf{Setting.} A cap rests on the sofa, and a coat rack with hooks is elsewhere in the room.\par
\textbf{Task.} Grasp the cap, carry it to the rack, position it on a suitable hook, and release it.\par
\textbf{Completion.} Leave the cap hanging from and supported by a valid rack hook after release. Merely placing it near the rack is insufficient.\par
\par\noindent\begin{minipage}{\linewidth}
\includegraphics[width=\linewidth]{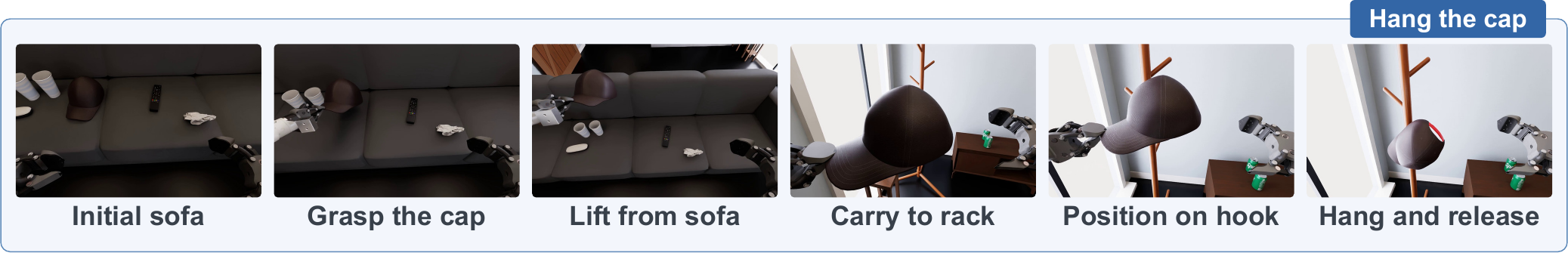}
\captionof{figure}{Procedure for the ``Hang the cap'' task.}\label{fig:task-06}
\end{minipage}\par
\end{RAADetail}

\begin{RAADetail}{7. Clear table waste}
\textbf{Request.} Clear the designated rubbish from the coffee table into the appropriate bins.\par
\textbf{Setting.} Designated waste items are on the coffee table, with waste and recycling bins nearby.\par
\textbf{Task.} Pick up each designated item and release it inside the appropriate waste or recycling bin. Items may be cleared in any order.\par
\textbf{Completion.} All designated items must be inside their correct bins and no longer held by the robot. Removing them from the table alone is insufficient.\par
\par\noindent\begin{minipage}{\linewidth}
\includegraphics[width=\linewidth]{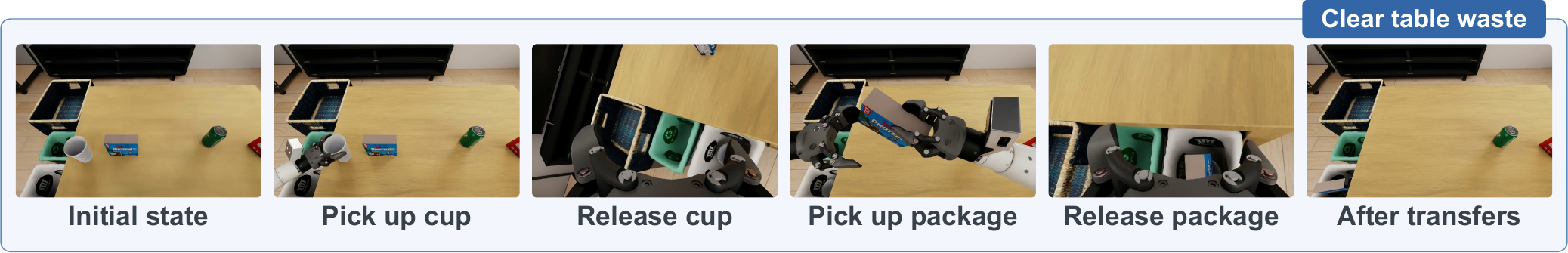}
\captionof{figure}{Procedure for the ``Clear table waste'' task.}\label{fig:task-07}
\end{minipage}\par
\end{RAADetail}

\begin{RAADetail}{8. Move food to the microwave}
\textbf{Request.} Move the designated food box from the refrigerator into the microwave and close both doors.\par
\textbf{Setting.} The target food box is inside the refrigerator among other food boxes. The refrigerator and microwave doors are initially closed.\par
\textbf{Task.} Open the refrigerator, retrieve the target box, and close the refrigerator door. Carry the box to the microwave, open its door, place and release the same box on the plate inside, and close the door. Heating is not required.\par
\textbf{Completion.} Complete the transfer and leave the same food box inside the microwave, with both appliance doors closed.\par
\par\noindent\begin{minipage}{\linewidth}
\includegraphics[width=\linewidth]{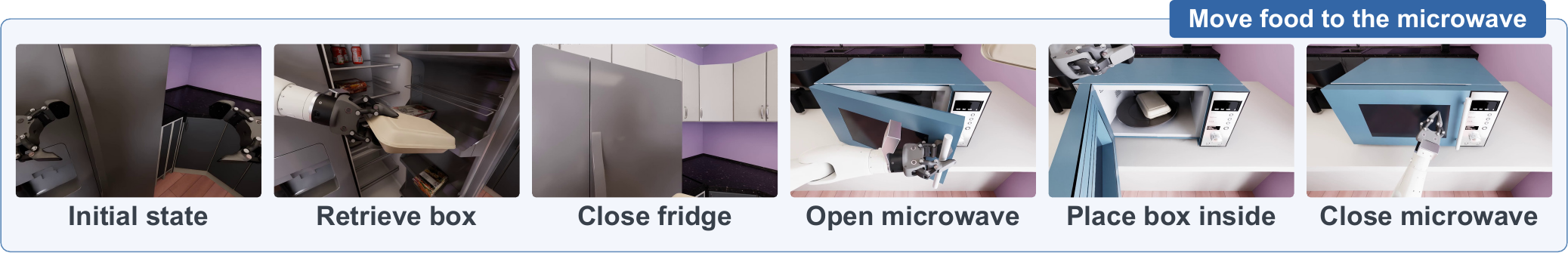}
\captionof{figure}{Procedure for the ``Move food to the microwave'' task.}\label{fig:task-08}
\end{minipage}\par
\end{RAADetail}

\begin{RAADetail}{9. Clear objects from the cloth}
\textbf{Request.} Move all designated objects off the tablecloth onto the permitted tabletop area.\par
\textbf{Setting.} The target objects initially rest on the tablecloth. A separate area on the same coffee table is designated for placing them.\par
\textbf{Task.} Move each designated object off the cloth and into the tabletop area specified by the instruction, in any order. Leave the cloth in place and keep other reserved tabletop areas clear.\par
\textbf{Completion.} All targets must be released and supported by the permitted tabletop area outside the cloth, with the cloth left in place.\par
\par\noindent\begin{minipage}{\linewidth}
\includegraphics[width=\linewidth]{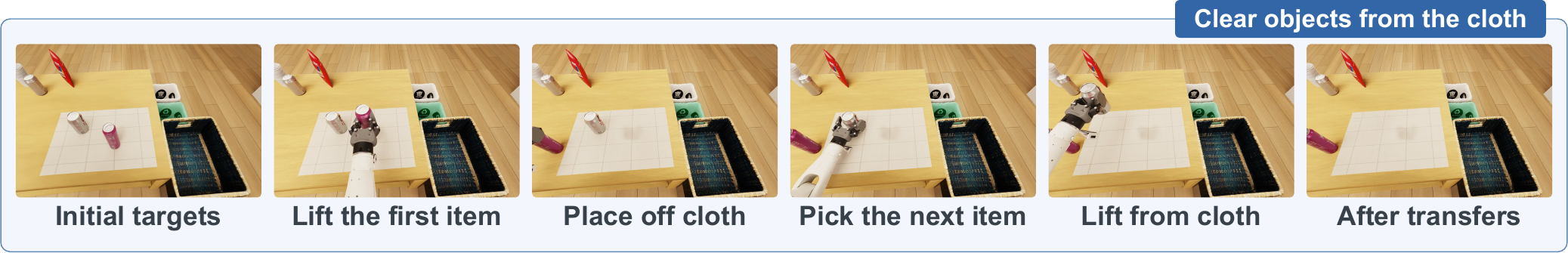}
\captionof{figure}{Procedure for the ``Clear objects from the cloth'' task.}\label{fig:task-09}
\end{minipage}\par
\end{RAADetail}

\begin{RAADetail}{10. Pack the bag}
\textbf{Request.} Put the requested personal object into the tote bag.\par
\textbf{Setting.} The target object rests at its specified source location. The destination tote bag is fixed in place and already open.\par
\textbf{Task.} Pick up the target, carry it to the tote bag, guide it through the opening, and release it inside. Opening or moving the bag is not part of the task.\par
\textbf{Completion.} Leave the same object released inside and supported by the bag, rather than on its rim or the surrounding surface.\par
\par\noindent\begin{minipage}{\linewidth}
\includegraphics[width=\linewidth]{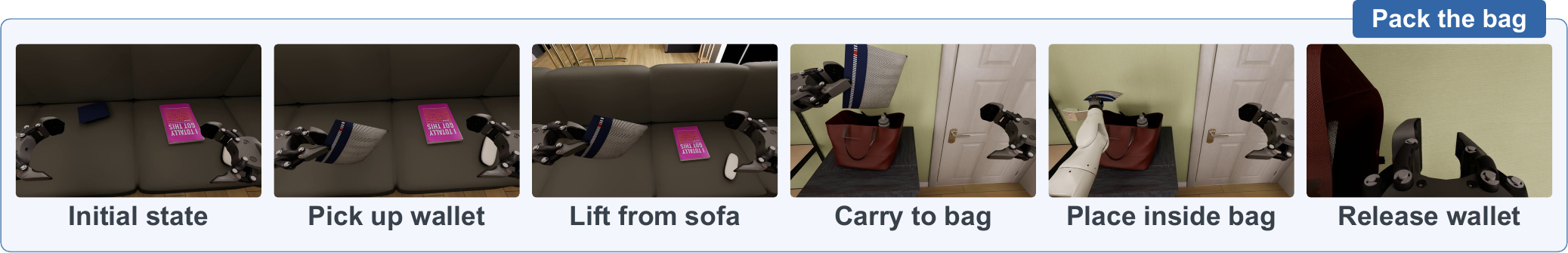}
\captionof{figure}{Procedure for the ``Pack the bag'' task.}\label{fig:task-10}
\end{minipage}\par
\end{RAADetail}

\begin{RAADetail}{11. Place cutlery on the tray}
\textbf{Request.} Place the designated utensils neatly on the large meal tray.\par
\textbf{Setting.} The requested utensils are in a cutlery holder beside the large meal tray.\par
\textbf{Task.} Grasp each designated utensil, extract it from its holder slot, carry it to the large meal tray, and release it inside. Follow the requested arrangement, such as placing the utensils neatly in parallel.\par
\textbf{Completion.} All requested utensils are placed on the tray in the specified arrangement.\par
\par\noindent\begin{minipage}{\linewidth}
\includegraphics[width=\linewidth]{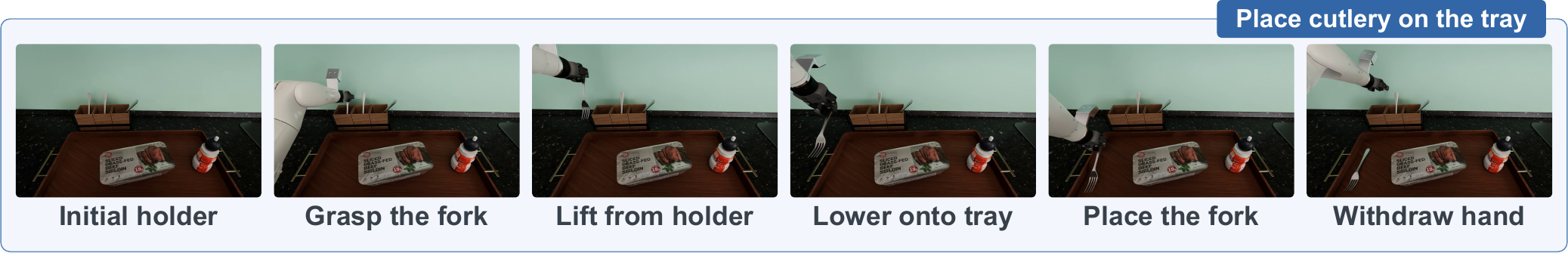}
\captionof{figure}{Procedure for the ``Place cutlery on the tray'' task.}\label{fig:task-11}
\end{minipage}\par
\end{RAADetail}

\begin{RAADetail}{12. Deliver the tray}
\textbf{Request.} Carry the large meal tray to the coffee table.\par
\textbf{Setting.} The large meal tray rests on its initial support and may be empty or already loaded with food boxes, cups, or utensils. The destination area on the coffee table is available.\par
\textbf{Task.} Grasp and lift the tray, carry it steadily to the designated area on the coffee table, lower it onto the tabletop, and release it. Preparing food or loading the tray beforehand is not part of this transport task.\par
\textbf{Completion.} Place and release the tray on the designated area of the coffee table, keeping any required contents on the tray.\par
\par\noindent\begin{minipage}{\linewidth}
\includegraphics[width=\linewidth]{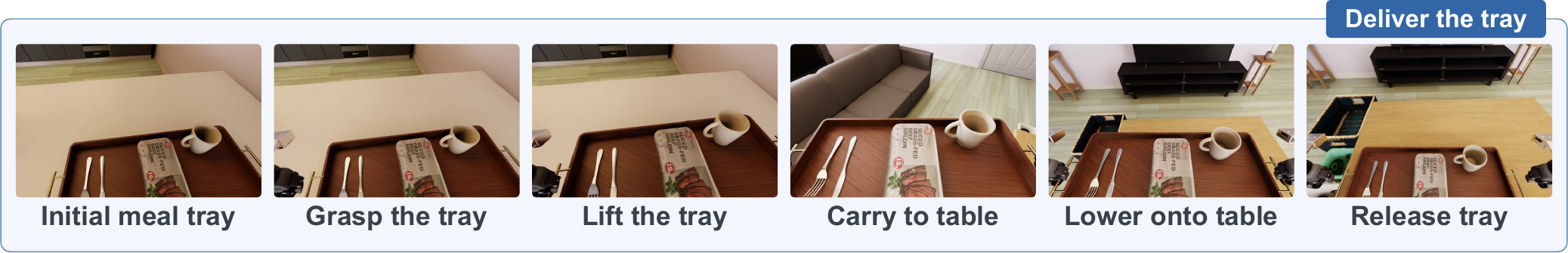}
\captionof{figure}{Procedure for the ``Deliver the tray'' task.}\label{fig:task-12}
\end{minipage}\par
\end{RAADetail}

\begin{RAADetail}{13. Move sofa items to the table}
\textbf{Request.} Move the requested object from the sofa to the coffee table.\par
\textbf{Setting.} A designated object, such as a book, remote control, wallet, or sunglasses case, rests on the sofa. The destination is the permitted object area on the coffee table.\par
\textbf{Task.} Pick up the requested object from the sofa, carry it to the coffee table, and release it in the permitted object area, keeping clear of the tablecloth and fixed wooden tray.\par
\textbf{Completion.} Leave the same object released in the permitted area and supported by the coffee-table surface.\par
\par\noindent\begin{minipage}{\linewidth}
\includegraphics[width=\linewidth]{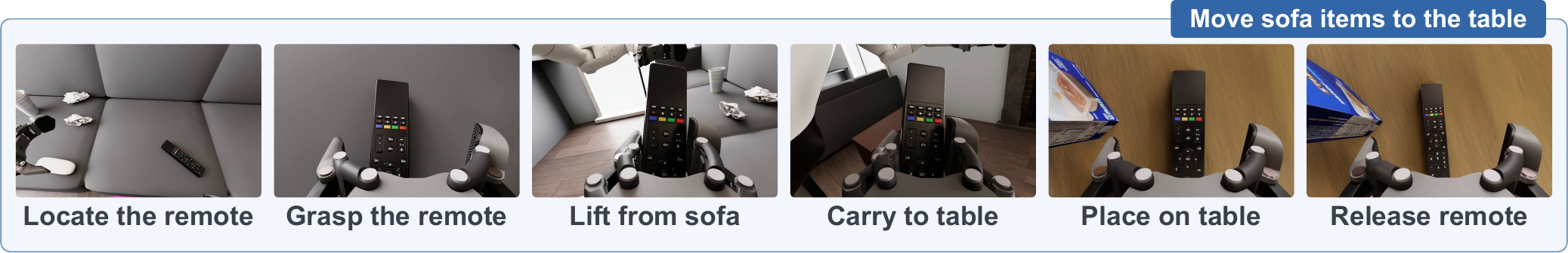}
\captionof{figure}{Procedure for the ``Move sofa items to the table'' task.}\label{fig:task-13}
\end{minipage}\par
\end{RAADetail}

\begin{RAADetail}{14. Store items in the basket}
\textbf{Request.} Put the requested tabletop object into the item-storage basket.\par
\textbf{Setting.} A wallet or sunglasses case rests on the specified table. The destination is the item-storage basket beside the waste bins.\par
\textbf{Task.} Pick up the target from the specified table, carry it to the item-storage basket, guide it through the opening, and release it inside. The tote bag and waste bins are not valid destinations.\par
\textbf{Completion.} Leave the same object released inside and supported by the designated item-storage basket.\par
\par\noindent\begin{minipage}{\linewidth}
\includegraphics[width=\linewidth]{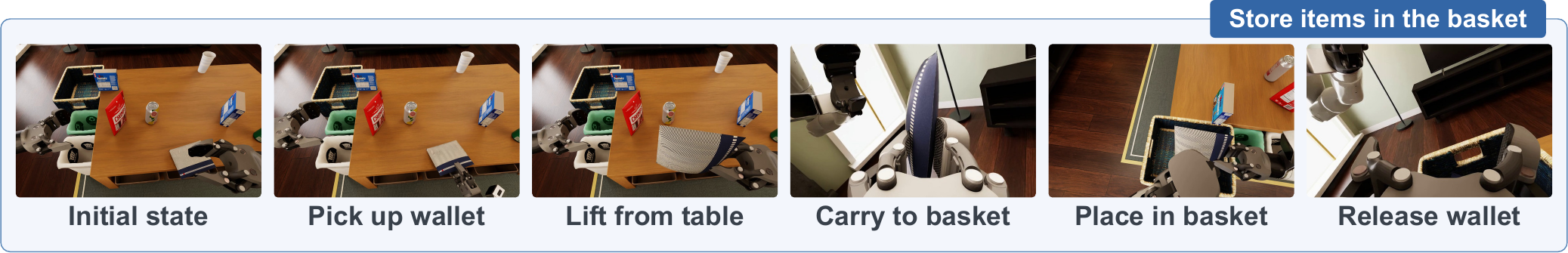}
\captionof{figure}{Procedure for the ``Store items in the basket'' task.}\label{fig:task-14}
\end{minipage}\par
\end{RAADetail}

\begin{RAADetail}{15. Heat food and retrieve the plate}
\textbf{Request.} Heat the food in the microwave, then take out the plate.\par
\textbf{Setting.} A food box rests on a plate inside the closed microwave. The microwave is initially off.\par
\textbf{Task.} Turn the control knob to start heating and wait for the light to turn off. Open the door, grasp and remove the plate while keeping the food box on it, and close the door. Take out the plate and food together rather than removing only the box.\par
\textbf{Completion.} Finish heating, retrieve the plate with the food box remaining on it, and close the microwave door.\par
\par\noindent\begin{minipage}{\linewidth}
\includegraphics[width=\linewidth]{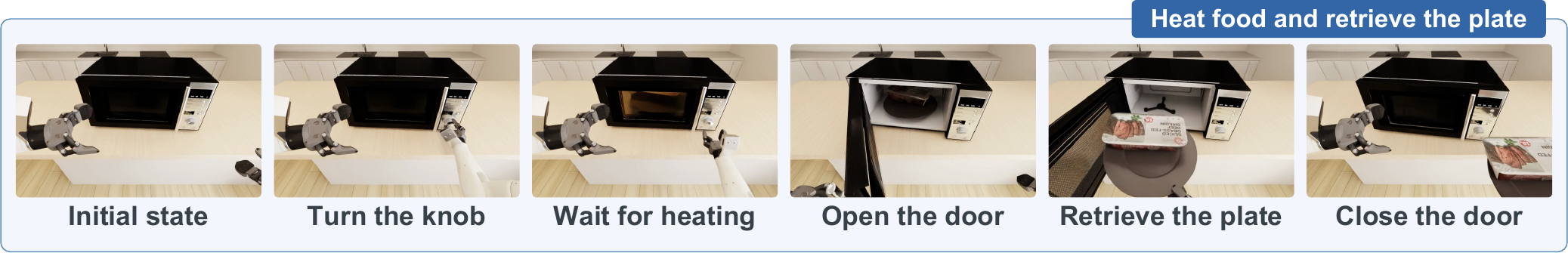}
\captionof{figure}{Procedure for the ``Heat food and retrieve the plate'' task, illustrated using atomic-action demonstrations.}\label{fig:task-15}
\end{minipage}\par
\end{RAADetail}

\begin{RAADetail}{16. Wipe the table}
\textbf{Request.} Pick up the brush, clean the stains on the coffee table, and return the brush to its original place.\par
\textbf{Setting.} A brush and one or more stains are on the coffee table. The brush is initially resting on the tabletop.\par
\textbf{Task.} Grasp and lift the brush, then wipe across the stained areas with the brush face in contact with the tabletop. After cleaning, return the brush to its original position and release it.\par
\textbf{Completion.} Remove the stains and return the brush to its original place.\par
\par\noindent\begin{minipage}{\linewidth}
\includegraphics[width=\linewidth]{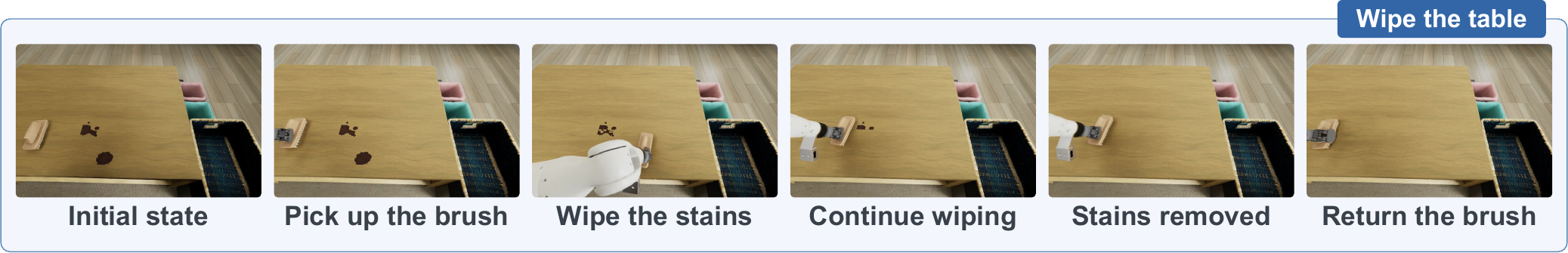}
\captionof{figure}{Procedure for the ``Wipe the table'' task.}\label{fig:task-16}
\end{minipage}\par
\end{RAADetail}

\begin{RAADetail}{17. Spread the tablecloth}
\textbf{Request.} Take the tablecloth from the cloth basket and spread it on the coffee table.\par
\textbf{Setting.} A tablecloth is stored in the designated cloth basket beside the coffee table. The intended tabletop region is clear of objects.\par
\textbf{Task.} Grasp the cloth, lift it out of the basket, and unfold it above the table. Lower and spread it over the designated tabletop region, adjust its position as needed, and release it.\par
\textbf{Completion.} Leave the tablecloth unfolded and spread over the designated tabletop area.\par
\par\noindent\begin{minipage}{\linewidth}
\centering
\includegraphics[width=\linewidth]{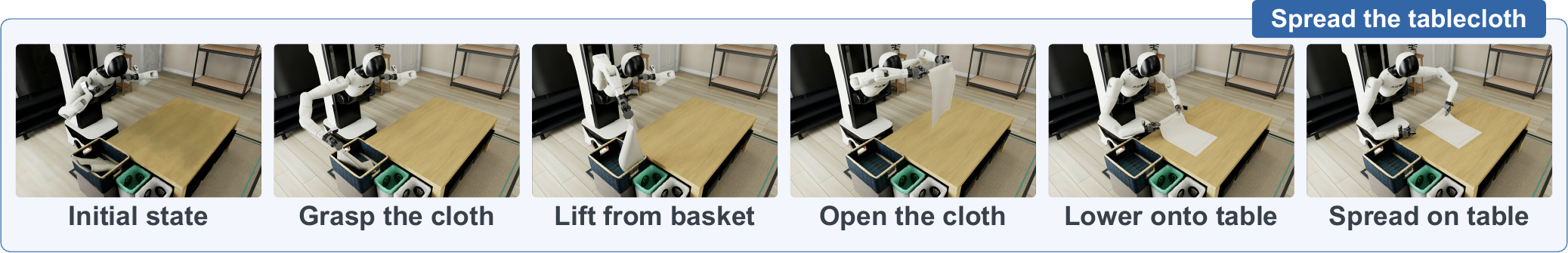}
\captionof{figure}{Procedure for the ``Spread the tablecloth'' task.}\label{fig:task-17}
\end{minipage}\par
\end{RAADetail}

\begin{RAADetail}{18. Collect the tablecloth}
\textbf{Request.} Collect the tablecloth from the coffee table and put it in the cloth basket.\par
\textbf{Setting.} A tablecloth lies on the coffee table, with no objects resting on it. The designated cloth basket is beside the table.\par
\textbf{Task.} Grasp an edge of the cloth, gather it, and lift the entire cloth clear of the tabletop. Move it over the designated cloth basket, lower it inside, and release it.\par
\textbf{Completion.} Place the entire tablecloth inside the designated basket, leaving no part on the table.\par
\par\noindent\begin{minipage}{\linewidth}
\centering
\includegraphics[width=\linewidth]{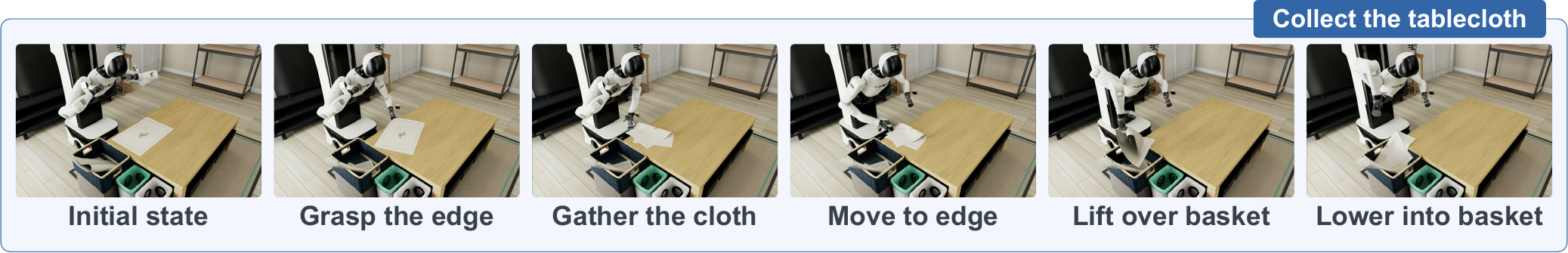}
\captionof{figure}{Procedure for the ``Collect the tablecloth'' task.}\label{fig:task-18}
\end{minipage}\par
\end{RAADetail}

\par\medskip\noindent\textbf{Long tasks}\par\nopagebreak[4]
\begin{RAADetail}{1. Stock the tray (Multi)}
\textbf{Request.} Bring all requested snacks and drinks from the shelf to the small wooden tray.\par
\textbf{Setting.} Multiple products are requested by type and quantity. They are on the shelf, and the small wooden tray is on the coffee table.\par
\textbf{Task.} Select the requested products, carry them to the coffee table, and release them inside the small wooden tray. Track the remaining items and avoid displacing products already delivered. The large meal tray is not the destination.\par
\textbf{Completion.} Place all requested products in the small wooden tray.\par
\par\noindent\begin{minipage}{\linewidth}
\includegraphics[width=\linewidth]{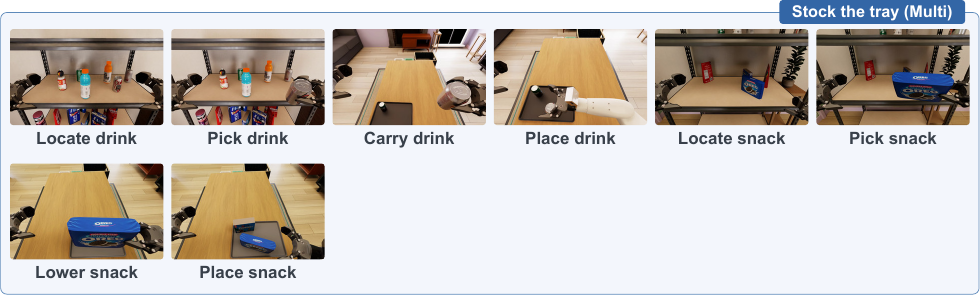}
\captionof{figure}{Drink and snack delivery components of the ``Stock the tray (Multi)'' task, illustrated using separate demonstrations.}\label{fig:long-task-01}
\end{minipage}\par
\end{RAADetail}

\begin{RAADetail}{2. Tidy the sofa area}
\textbf{Request.} Put the book on the coffee table, store the wallet in the coffee-table drawer and close it, and hang the cap on the coat rack.\par
\textbf{Setting.} A book, a wallet, and a cap are on the sofa. The coffee table, its initially closed drawer, and the coat rack are available.\par
\textbf{Task.} Transfer the book to the designated coffee-table surface. Open the drawer, place and release the wallet inside, and close the drawer. Hang and release the cap on a valid coat-rack hook. Complete these transfers without undoing earlier placements.\par
\textbf{Completion.} The book rests on the designated table surface, the wallet remains inside the closed drawer, and the cap is supported by the coat-rack hook after release. All three conditions must remain satisfied at the end.\par
\par\noindent\begin{minipage}{\linewidth}
\includegraphics[width=\linewidth]{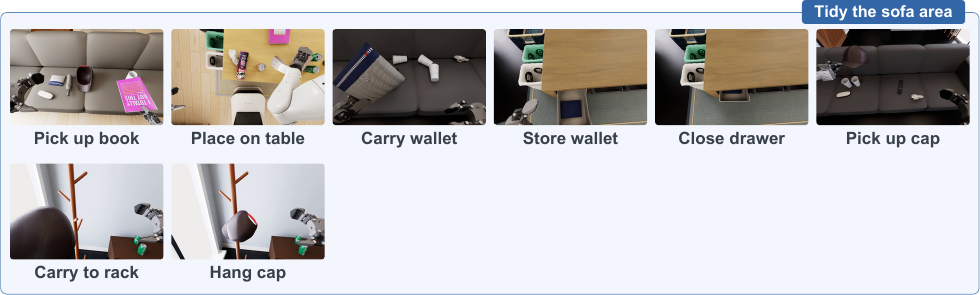}
\captionof{figure}{Book transfer, wallet storage, and cap hanging for the ``Tidy the sofa area'' task, illustrated using separate demonstrations.}\label{fig:long-task-02}
\end{minipage}\par
\end{RAADetail}

\begin{RAADetail}{3. Tidy the table}
\textbf{Request.} Put the designated table waste in the correct bins and the requested belongings in the item-storage basket.\par
\textbf{Setting.} Specified waste items and reusable belongings are on the table. The waste bins and the item-storage basket are available.\par
\textbf{Task.} Identify the requested waste items and dispose of each in the assigned bin. Pick up the specified wallet or sunglasses case from the table and place it in the item-storage basket. Do not discard reusable belongings or use the tote bag as the storage destination. Preserve earlier placements while completing the remaining transfers.\par
\textbf{Completion.} Every designated waste item is in its correct bin, and every requested belonging is inside and supported by the item-storage basket. These conditions must remain satisfied at the end.\par
\par\noindent\begin{minipage}{\linewidth}
\includegraphics[width=\linewidth]{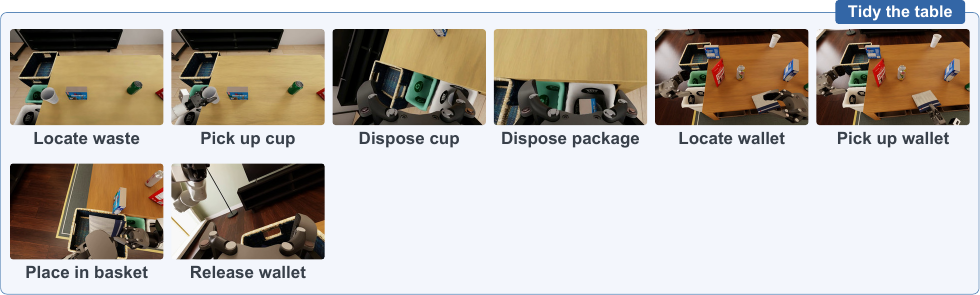}
\captionof{figure}{Waste disposal and belonging storage for the ``Tidy the table'' task, illustrated using separate demonstrations.}\label{fig:long-task-03}
\end{minipage}\par
\end{RAADetail}

\begin{RAADetail}{4. Replace the tablecloth}
\textbf{Request.} Replace the used tablecloth with a clean one.\par
\textbf{Setting.} The used cloth covers the table. A clean cloth and the cloth basket are available from the start.\par
\textbf{Task.} Clear obstructions when needed, remove the used cloth, fold it, and put it away. Retrieve the clean cloth, spread it over the intended tabletop region, and release it.\par
\textbf{Completion.} The used cloth is folded and stored, and the clean cloth is released and spread over the intended tabletop region.\par
\par\noindent\begin{minipage}{\linewidth}
\includegraphics[width=\linewidth]{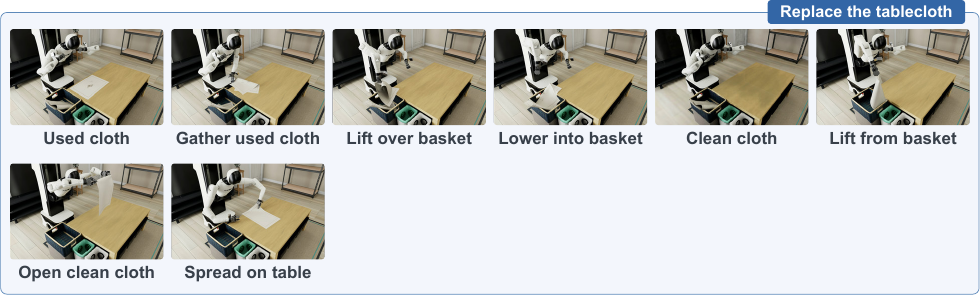}
\captionof{figure}{Cloth collection and spreading components of the ``Replace the tablecloth'' task, illustrated using separate demonstrations.}\label{fig:long-task-04}
\end{minipage}\par
\end{RAADetail}

\begin{RAADetail}{5. Find and pack an item}
\textbf{Request.} Find the requested object and put it in the tote bag.\par
\textbf{Setting.} A wallet or sunglasses case is placed somewhere in the room before the episode begins. The instruction identifies the target but does not reveal its location. The destination is the designated open tote bag.\par
\textbf{Task.} Search the room, locate the requested object, pick it up, and carry it to the bag. Guide the same object through the opening and release it inside. Search and packing take place in the same scene without resetting the object.\par
\textbf{Completion.} Locate the requested object, complete its pickup and packing sequence, and leave it inside and supported by the designated bag.\par
\par\noindent\begin{minipage}{\linewidth}
\includegraphics[width=\linewidth]{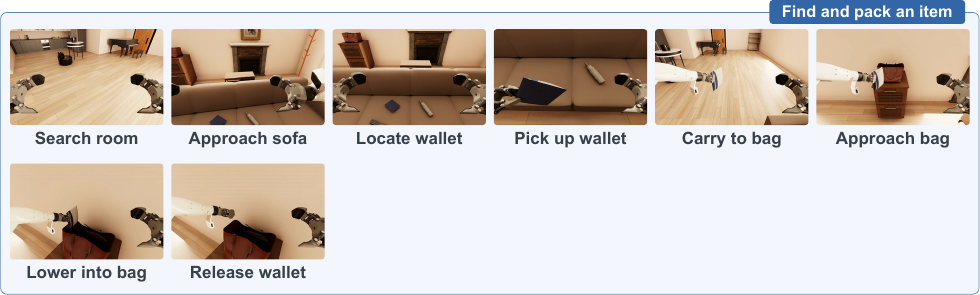}
\captionof{figure}{Search and packing of a wallet, shown as an excerpt from one continuous recording.}\label{fig:long-task-05}
\end{minipage}\par
\end{RAADetail}

\begin{RAADetail}{6. Prepare a meal}
\textbf{Request.} Take the requested food from the refrigerator, heat it in the microwave, and place it with the requested cutlery on the large meal tray.\par
\textbf{Setting.} The requested food box is in the refrigerator. A microwave, a large meal tray, and the requested cutlery are available.\par
\textbf{Task.} Retrieve the food box and close the refrigerator. Heat the food in the microwave, retrieve it, and close the microwave door. Transfer the heated food box from the microwave plate to the large meal tray, and add the requested cutlery. Leave the microwave plate off the tray.\par
\textbf{Completion.} The heated food box and all requested cutlery are on the large meal tray. Delivering the tray is not required.\par
\par\noindent\begin{minipage}{\linewidth}
\includegraphics[width=\linewidth]{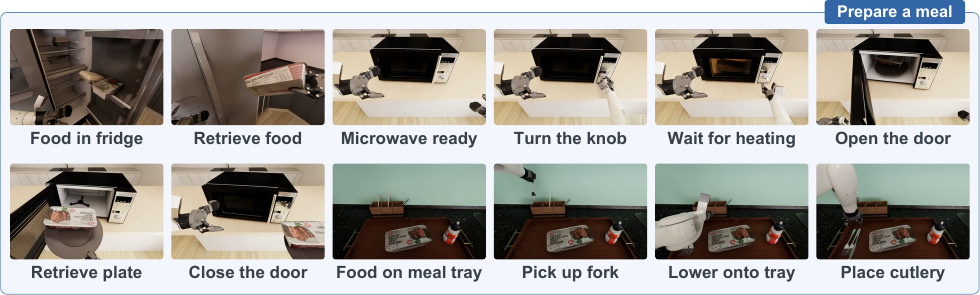}
\captionof{figure}{Refrigerator, microwave, and cutlery components of the ``Prepare a meal'' task, illustrated using separate demonstrations.}\label{fig:long-task-06}
\end{minipage}\par
\end{RAADetail}

\begin{RAADetail}{7. Prepare coffee and pastry}
\textbf{Request.} Prepare a cup of coffee and load and retrieve the designated pastry using the air fryer.\par
\textbf{Setting.} A cup and coffee beans are available at the coffee machine. The target pastry is on a plate beside the air fryer, whose basket is initially closed.\par
\textbf{Task.} Position the cup under the coffee outlet, pour the beans, return and release the bean bag, operate the machine, and retrieve the coffee cup. For the pastry, pull out the basket, place the target inside, and close it. Then remove the basket completely, set it down securely, and retrieve the same pastry. This task instance requires neither pastry heating nor returning the pastry to the plate.\par
\textbf{Completion.} Prepare the coffee and retrieve the cup, and complete the pastry loading and retrieval sequence. The cup and pastry do not need to be held simultaneously at the end.\par
\par\noindent\begin{minipage}{\linewidth}
\includegraphics[width=\linewidth]{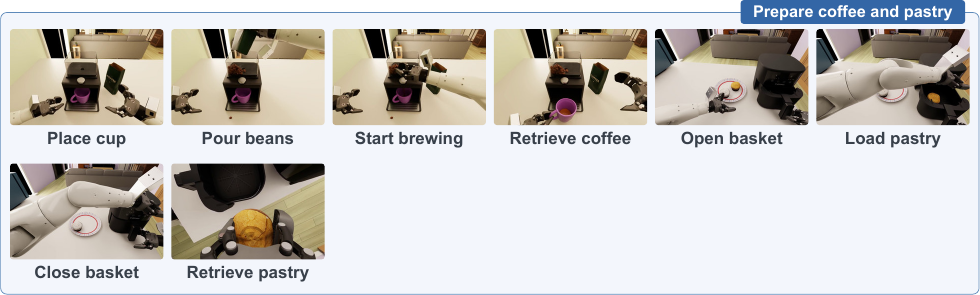}
\captionof{figure}{Coffee preparation and pastry retrieval for the ``Prepare coffee and pastry'' task, illustrated using separate demonstrations.}\label{fig:long-task-07}
\end{minipage}\par
\end{RAADetail}

\begin{RAADetail}{8. Pack bags and baskets}
\textbf{Request.} Collect the listed objects and put each in its assigned bag or item-storage basket.\par
\textbf{Setting.} A wallet and a sunglasses case are placed at specified locations in the room, each assigned to a bag or item-storage basket.\par
\textbf{Task.} Pick up each requested object from its source, carry it to the assigned container, and release it inside. Keep track of the remaining objects and preserve earlier placements. The assignments may use the same container type for all targets or a mixture of bags and baskets.\par
\textbf{Completion.} Complete all requested transfers and leave every target inside and supported by its assigned container at the end.\par
\par\noindent\begin{minipage}{\linewidth}
\includegraphics[width=\linewidth]{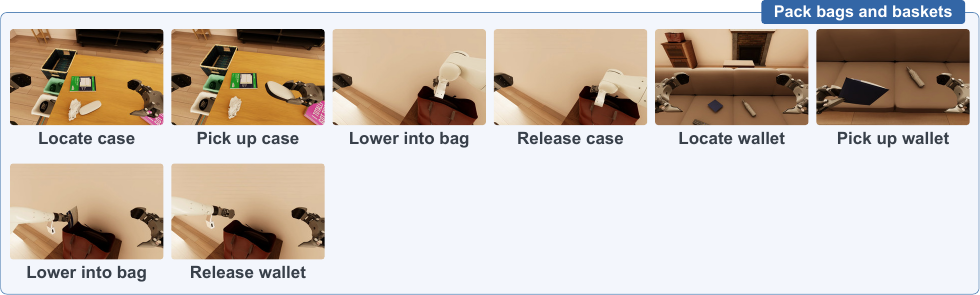}
\captionof{figure}{Successive packing of a glasses case and a wallet, shown as excerpts from one continuous recording.}\label{fig:long-task-08}
\end{minipage}\par
\end{RAADetail}

\subsection{Task Discovery}
\begin{RAADetail}{Identify task-relevant objects}
\textbf{Task overview.} We evaluate the agent's ability to infer which household tasks are needed and identify the objects involved across four scenarios: dinner preparation, hosting guests, preparing to leave, and room tidying.\par
\textbf{Input and setting.} The agent receives household memory, a high-level scenario request, and camera observations. It explores the home, infers which objects need to be handled, and indicates them by touching each with either gripper. The request does not explicitly list the target objects.\par
\textbf{Completion and scoring.} Performance is measured by object coverage: the number of distinct target objects touched divided by the total number of targets, $N$. Each episode allows contact with at most $N+2$ distinct physical objects, including targets and non-targets. Repeated contacts with the same object count once, and non-target contacts receive no credit. Only gripper-finger contacts count. The episode ends when all targets are touched, the touch budget is exhausted, or the time limit is reached. Grasping or completing the inferred household tasks is not required.\par
\par\noindent\begin{minipage}{\linewidth}
\includegraphics[width=\linewidth]{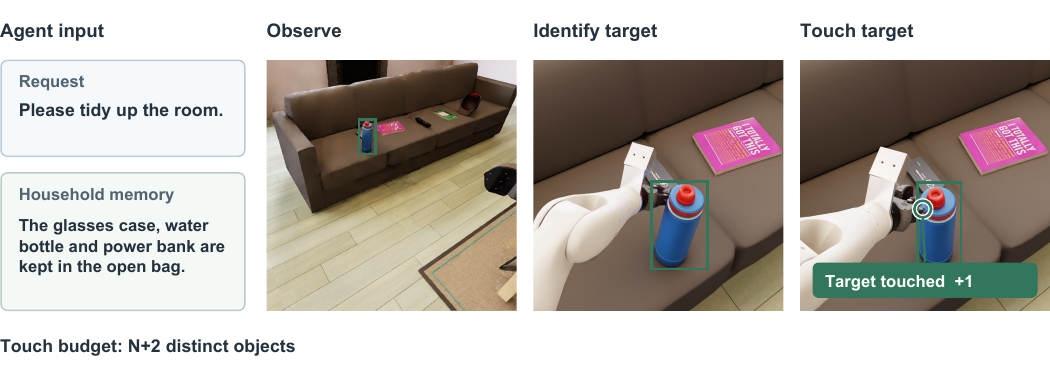}
\captionof{figure}{Task Discovery: identifying and touching a task-relevant object.}\label{fig:task-discovery}
\end{minipage}\par
\end{RAADetail}
\subsection{Context Reasoning}
\textbf{Task overview.} We evaluate the agent’s ability to infer which object meets a contextual requirement across four tasks: drink selection, snack selection, waste selection, and sofa object selection.\par
\textbf{Input and setting.} The agent receives a task instruction, contextual information, and camera observations. The context describes user preferences, intended uses, classification rules, or updated requests. The agent must use this information to identify the appropriate object among the visible candidates.\par
\textbf{Completion and scoring.} The agent selects an object by touching it with either gripper. Touching the correct candidate earns 1 point and ends the episode; touching an incorrect candidate earns 0 and ends the episode. Simultaneous correct and incorrect contacts also fail. Episodes that time out without selecting a candidate receive zero credit. Contacts with background furniture do not count as selections. Grasping, lifting, carrying, placing, and disposal are not required.\par

\begin{RAADetail}{1. Drink selection}
\textbf{Setting.} Several drinks are stocked on a shelf. The supplied situation specifies the user's requirements, such as flavour, sugar content, or caffeine preference, rather than a target package.\par
\textbf{Task.} Resolve the current request and select the available drink that satisfies its requirements. When the context contains competing or revised requests, follow the confirmed one.\par
\textbf{Example.} The illustrated request asks for sweet-tasting cola with zero sugar. Diet Coke is the correct choice; unsweetened sparkling water does not satisfy the cola requirement.\par
\par\noindent\begin{minipage}{\linewidth}
\centering
\includegraphics[width=\linewidth]{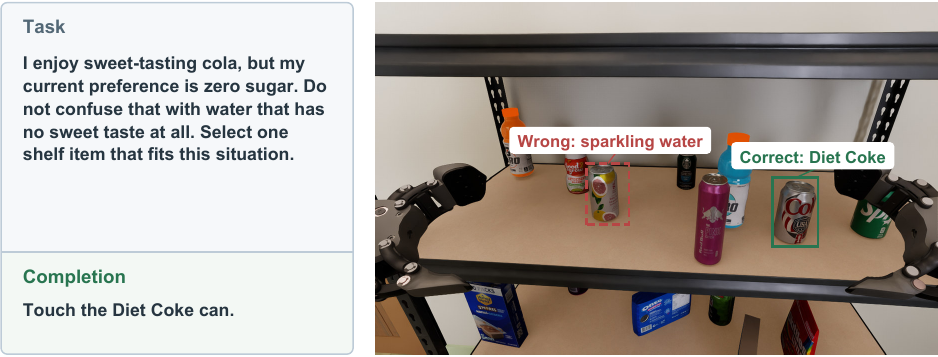}
\captionof{figure}{Context Reasoning: drink selection.}\label{fig:context-drink}
\end{minipage}\par
\end{RAADetail}

\begin{RAADetail}{2. Snack selection}
\textbf{Setting.} A shelf contains snacks with different flavours, textures, and ingredients, together with other stocked products. The situation describes the desired snack properties or the request that currently takes priority.\par
\textbf{Task.} Infer the requested properties and select the matching snack from the visible inventory. A broadly similar product is insufficient when it conflicts with a stated requirement.\par
\textbf{Example.} The illustrated request calls for a crunchy savoury snack with a smoky, grill-like flavour. Barbecue-flavoured crisps satisfy it; sour-cream-and-onion crisps do not match the requested flavour.\par
\par\noindent\begin{minipage}{\linewidth}
\centering
\includegraphics[width=\linewidth]{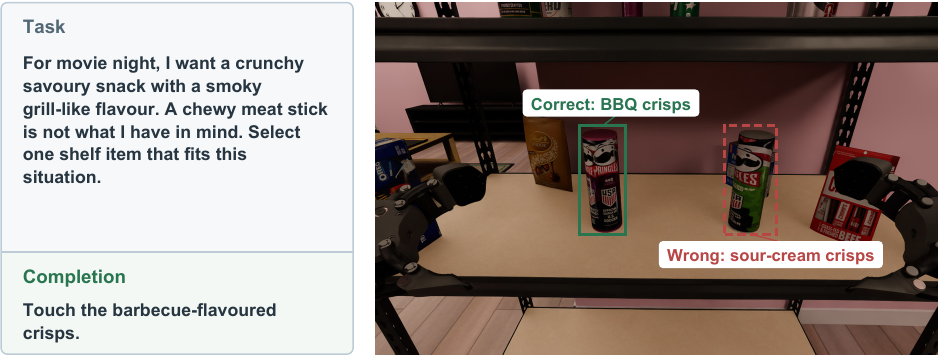}
\captionof{figure}{Context Reasoning: snack selection.}\label{fig:context-snack}
\end{minipage}\par
\end{RAADetail}

\begin{RAADetail}{3. Waste selection}
\textbf{Setting.} Several candidate waste objects are placed on the coffee table. The instruction requests one category, and the accompanying context explicitly defines the recyclable and non-recyclable categories for this evaluation.\par
\textbf{Task.} Apply the supplied classification rules and touch the sole candidate in the requested category. The task tests category-based selection, not transport to a bin.\par
\textbf{Example.} The illustrated instruction requests the only recyclable item. Under the supplied rules, the dry cardboard box is correct and the used toothbrush is incorrect.\par
\par\noindent\begin{minipage}{\linewidth}
\centering
\includegraphics[width=\linewidth]{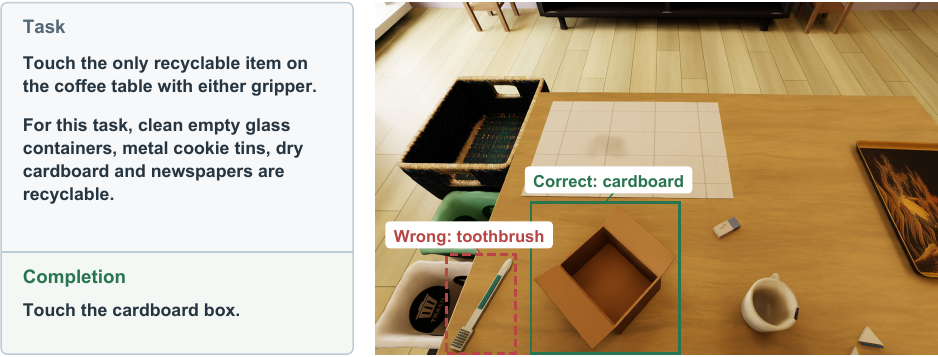}
\captionof{figure}{Context Reasoning: waste selection.}\label{fig:context-waste}
\end{minipage}\par
\end{RAADetail}

\begin{RAADetail}{4. Sofa object selection}
\textbf{Setting.} Eight personal objects are placed on the sofa: a TV remote, a wallet, a sunglasses case, a power bank, a cap, an umbrella, a water bottle, and a green notebook. The context supplies relevant object-use facts, and the scenario describes the user's current need.\par
\textbf{Task.} Infer which object serves that need and touch it. Follow the latest request when an earlier plan is cancelled or superseded; no sofa-to-table transfer is required.\par
\textbf{Example.} The illustrated user needs rain protection over the shoulders, not only the top of the head. The umbrella is correct; the power bank cannot meet this need.\par
\par\noindent\begin{minipage}{\linewidth}
\centering
\includegraphics[width=\linewidth]{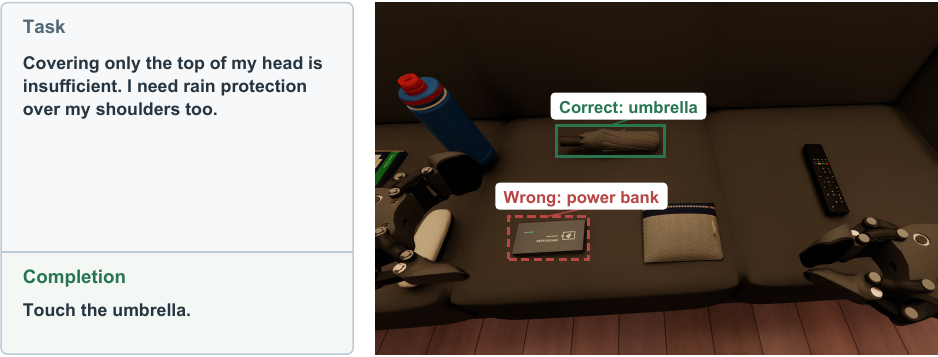}
\captionof{figure}{Context Reasoning: sofa object selection.}\label{fig:context-sofa}
\end{minipage}\par
\end{RAADetail}

\subsection{Failure Recovery}
\textbf{Task overview.} We evaluate the agent's ability to assess a failed action state, adjust its actions, and complete the task. The five settings cover missed grasps of a shelf drink, a coffee-bean bag, a TV remote, and a used cup, together with a tipped-cup state.\par
\textbf{Input and setting.} The agent receives the task instruction, camera observations, and robot-state feedback. Each episode starts from a predefined failed state, with the target still in the scene but not held. The figures pair a normal-operation reference (left) with a recovery initial state (right), using separate recordings.\par
\textbf{Completion and scoring.} Performance is measured using task-specific process scores during recovery.\par

\begin{RAADetail}{1. Shelf pickup (Missed grasp)}
\textbf{Setting.} The requested drink remains upright on the shelf after a missed grasp, with an empty gripper nearby.\par
\textbf{Task.} Adjust the grasp, lift the same drink clear of the shelf, and hold it steadily for three seconds. Tray delivery is not required.\par
\textbf{Example.} The illustrated target is a juice bottle. The left image shows it held by the gripper; in the right image, it remains on the shelf beside the empty hand.\par
\par\noindent\begin{minipage}{\linewidth}
\centering
\includegraphics[width=\linewidth]{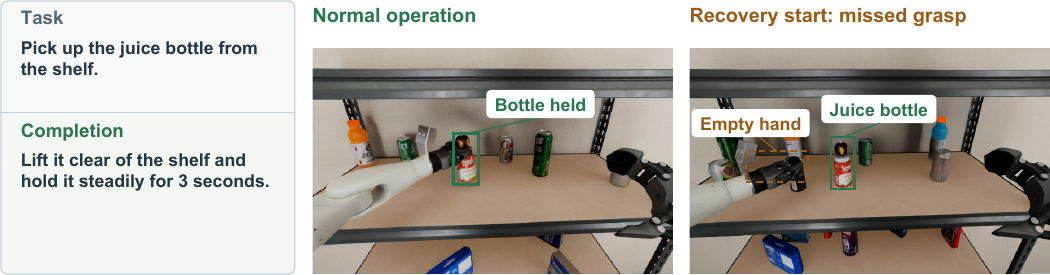}
\captionof{figure}{Failure Recovery: normal shelf pickup (left) and a missed-grasp state (right).}\label{fig:recovery-shelf}
\end{minipage}\par
\end{RAADetail}

\begin{RAADetail}{2. Pour coffee beans (Missed grasp)}
\textbf{Setting.} The coffee-bean bag remains upright on the counter beside the coffee machine after a missed grasp, with an empty gripper nearby.\par
\textbf{Task.} Adjust the grasp, pick up the bag, pour the beans into the hopper, and return the bag upright to the counter. Preparing or retrieving a cup of coffee is not required.\par
\textbf{Example.} The left image shows the bag held during normal pickup. In the right image, the empty gripper is beside the bag, which remains on the counter.\par
\par\noindent\begin{minipage}{\linewidth}
\centering
\includegraphics[width=\linewidth]{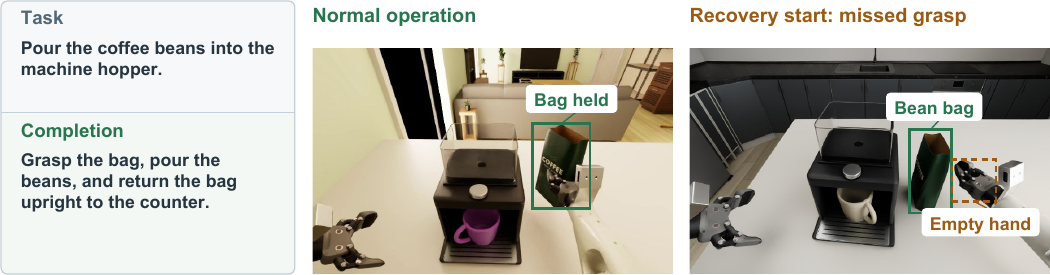}
\captionof{figure}{Failure Recovery: normal bean-bag pickup (left) and a missed-grasp state (right).}\label{fig:recovery-coffee-bag}
\end{minipage}\par
\end{RAADetail}

\begin{RAADetail}{3. Move sofa items to the table (Missed grasp)}
\textbf{Setting.} The TV remote remains in the sofa area after a missed grasp, with an empty gripper nearby. The coffee table is the destination.\par
\textbf{Task.} Adjust the grasp, pick up the same remote, and release it fully within and supported by the coffee tabletop. This recovery variant accepts any valid position on that tabletop; the standard sofa-to-table task specifies a restricted placement area.\par
\textbf{Example.} The left image shows the remote held after a normal pickup. In the right image, the remote remains in the sofa area and the nearby gripper is empty.\par
\par\noindent\begin{minipage}{\linewidth}
\centering
\includegraphics[width=\linewidth]{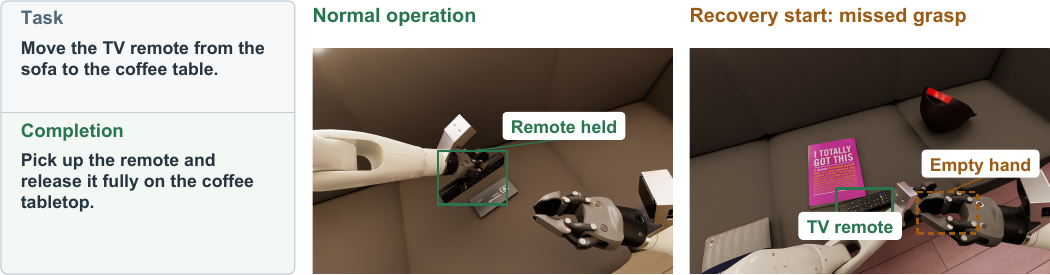}
\captionof{figure}{Failure Recovery: normal TV-remote pickup (left) and a missed-grasp state (right).}\label{fig:recovery-sofa}
\end{minipage}\par
\end{RAADetail}

\begin{RAADetail}{4. Clear table waste (Missed grasp)}
\textbf{Setting.} The used cup remains upright on the coffee table after a missed grasp, with an empty gripper nearby.\par
\textbf{Task.} Adjust the grasp, pick up the same cup, and release it inside its assigned waste bin.\par
\textbf{Example.} The left image shows a normal cup pickup. In the right image, the cup remains upright on the table beside the empty gripper.\par
\par\noindent\begin{minipage}{\linewidth}
\centering
\includegraphics[width=\linewidth]{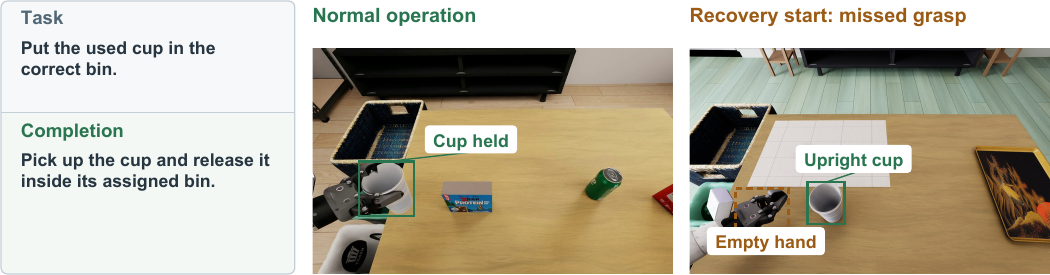}
\captionof{figure}{Failure Recovery: normal cup pickup (left) and a missed-grasp state (right).}\label{fig:recovery-cup-missed}
\end{minipage}\par
\end{RAADetail}

\begin{RAADetail}{5. Clear table waste (Tipped cup)}
\textbf{Setting.} The used cup lies on its side on the coffee table and is not held by either gripper.\par
\textbf{Task.} Adapt the pickup to the cup's current position and orientation, then release the same cup inside its assigned waste bin.\par
\textbf{Example.} The left image shows a normal cup handover between the two grippers. The right image shows the tipped cup on the table, with both grippers empty.\par
\par\noindent\begin{minipage}{\linewidth}
\centering
\includegraphics[width=\linewidth]{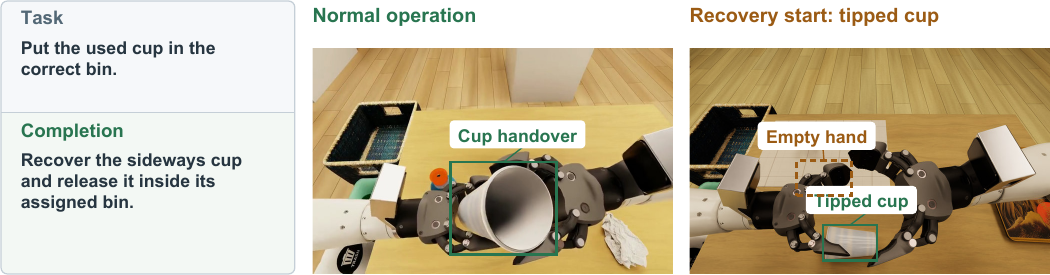}
\captionof{figure}{Failure Recovery: normal cup handover (left) and a tipped-cup state (right).}\label{fig:recovery-cup-tipped}
\end{minipage}\par
\end{RAADetail}
\subsection{Robustness}
\textbf{Task overview.} We evaluate the agent's ability to adapt to disturbances during manipulation. The six settings cover shelf-target relocation, shelf lighting changes, coffee-bag relocation, coffee-machine displacement, combined bag and machine movement, and lighting changes during button operation.\par
\textbf{Input and setting.} The agent receives the task instruction, camera observations, and robot-state feedback. During execution, a human hand moves a task object or the scene illumination changes. The agent must respond to the changed conditions while continuing the same task.\par
\textbf{Completion and scoring.} Performance is measured using task-specific process scores.\par

\begin{RAADetail}{1. Shelf pickup (Relocated)}
\textbf{Setting.} One product on the shelf is designated as the target. As the robot approaches, a visible human hand moves the target to another position on the same shelf level and releases it.\par
\textbf{Task.} Locate the product at its new position, grasp it, and lift it clear of the shelf. Tray placement is not required.\par
\textbf{Example.} The hand moves the target bottle away from the robot's planned grasp position. The agent must update its approach to pick up the same bottle.\par
\par\noindent\begin{minipage}{\linewidth}
\centering
\includegraphics[width=\linewidth]{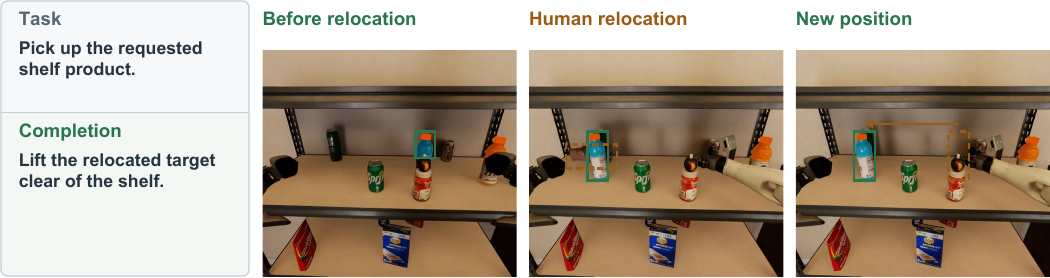}
\captionof{figure}{Robustness: shelf-target relocation by a human hand.}\label{fig:robustness-shelf-move}
\end{minipage}\par
\end{RAADetail}
\begin{RAADetail}{2. Shelf pickup (Lighting)}
\textbf{Setting.} A single requested product is on the shelf. Light intensity alternates between its original value and one-quarter of that value at intervals of two to three seconds. Object positions remain unchanged.\par
\textbf{Task.} Locate and pick up the requested product under the changing illumination, lifting it clear of the shelf and holding it steadily for three seconds. Tray placement is not required.\par
\textbf{Example.} The figure shows the same shelf under normal and reduced illumination. The target product remains in place, but its appearance changes with the lighting.\par
\par\noindent\begin{minipage}{\linewidth}
\centering
\includegraphics[width=\linewidth]{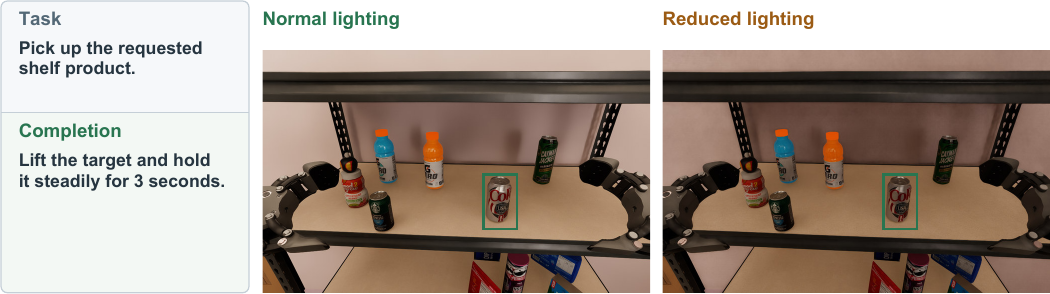}
\captionof{figure}{Robustness: the same shelf under normal and reduced illumination.}\label{fig:robustness-shelf-lighting}
\end{minipage}\par
\end{RAADetail}
\begin{RAADetail}{3. Pour coffee beans (Bag moved)}
\textbf{Setting.} The coffee-bean bag is upright on the counter beside the machine. As the robot approaches to grasp it, a human hand moves the bag to another position on the counter and releases it.\par
\textbf{Task.} Locate and grasp the relocated bag, pour the beans into the hopper, and return the bag upright to the counter. Preparing or retrieving a cup of coffee is not required.\par
\textbf{Example.} The bag moves before the gripper reaches it. The agent must adjust the pickup to the new bag position before continuing with the pour.\par
\par\noindent\begin{minipage}{\linewidth}
\centering
\includegraphics[width=\linewidth]{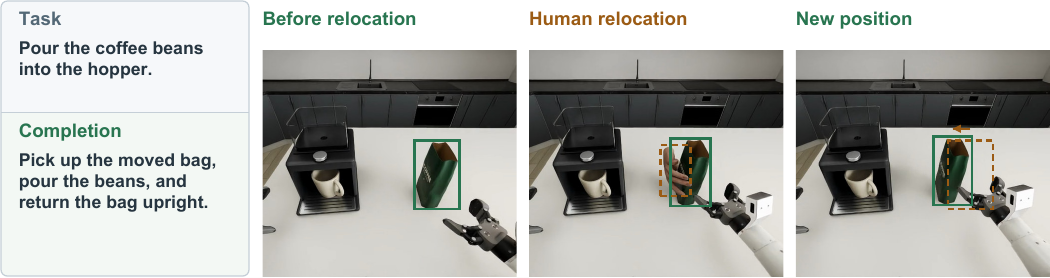}
\captionof{figure}{Robustness: coffee-bag relocation before pickup.}\label{fig:robustness-coffee-bag}
\end{minipage}\par
\end{RAADetail}

\begin{RAADetail}{4. Pour coffee beans (Machine moved)}
\textbf{Setting.} The robot holds the coffee-bean bag above the hopper. A human hand pushes the coffee machine while the bag is tilted for pouring.\par
\textbf{Task.} Track the displaced hopper, adjust the bag's position to continue pouring into it, and return the bag upright to the counter.\par
\textbf{Example.} The hopper moves away from the bag opening during the pour. The agent must realign the bag with the new hopper position rather than continue pouring at the old one.\par
\par\noindent\begin{minipage}{\linewidth}
\centering
\includegraphics[width=\linewidth]{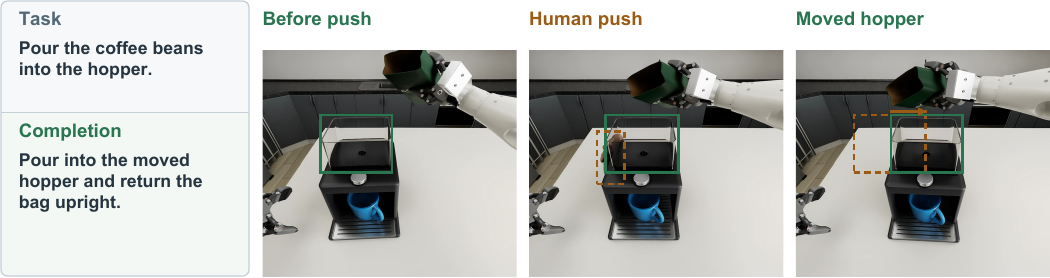}
\captionof{figure}{Robustness: coffee-machine displacement during pouring.}\label{fig:robustness-coffee-machine}
\end{minipage}\par
\end{RAADetail}

\begin{RAADetail}{5. Pour coffee beans (Both moved)}
\textbf{Setting.} Two disturbances occur in sequence: a human hand first relocates the bag before pickup, then pushes the coffee machine during pouring.\par
\textbf{Task.} Adjust the pickup to the relocated bag, realign the pour after the machine moves, and return the bag upright to the counter.\par
\textbf{Example.} Finding the moved bag resolves the first disturbance, but the agent must update its actions again when the hopper shifts during pouring.\par
\par\noindent\begin{minipage}{\linewidth}
\centering
\includegraphics[width=\linewidth]{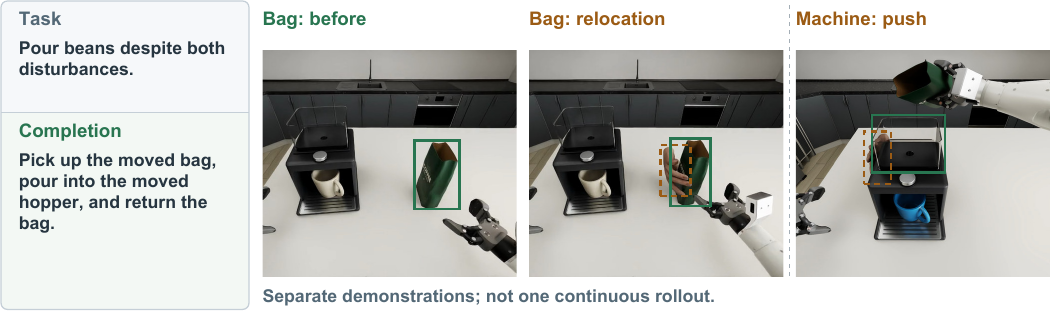}
\captionof{figure}{Robustness: bag relocation and machine displacement, illustrated using separate demonstrations.}\label{fig:robustness-coffee-combined}
\end{minipage}\par
\end{RAADetail}

\begin{RAADetail}{6. Operate the coffee-machine button (Lighting)}
\textbf{Setting.} The coffee machine and cup are ready for dispensing. Illumination changes during button operation, including changes in brightness, side lighting, and brief flashes.\par
\textbf{Task.} Locate and press the machine button to start dispensing, then press it again to stop after collecting the required amount of coffee.\par
\textbf{Example.} Side lighting or a brief flash changes the button's appearance. The agent must still identify and operate the same control to start and stop dispensing.\par
\par\noindent\begin{minipage}{\linewidth}
\centering
\includegraphics[width=\linewidth]{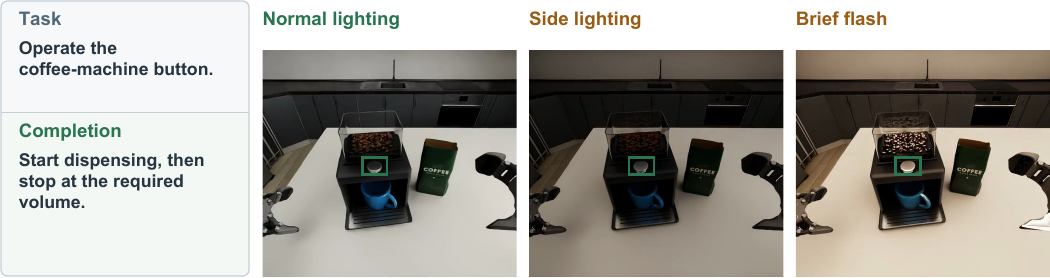}
\captionof{figure}{Robustness: the coffee-machine button under normal lighting, side lighting, and a brief flash.}\label{fig:robustness-button-lighting}
\end{minipage}\par
\end{RAADetail}

\subsection{Open-ended Scenarios}
\textbf{Task overview.} We evaluate the agent's ability to infer, plan, and execute household tasks from a high-level request in four scenarios: preparing dinner, hosting guests, preparing to leave, and tidying the room. Each episode combines multiple subtasks that must be completed while preserving earlier results.\par

\textbf{Input and setting.} Each room is generated from a selected layout and a scenario-specific task specification defining the required objects, appliances, source locations, destinations, and initial states. All subtasks share the same physical scene and object instances, without resets between subtasks. The agent receives household memory, a scenario request, camera observations, and robot-state feedback, but not the underlying subtask list or target bindings. Household memory describes personal preferences, routines, and storage arrangements. The agent combines this information with scene observations to identify the required tasks and plan their execution, respecting dependencies such as preparing food before serving it.\par

\textbf{Completion and scoring.} Native subtask scores reward verified physical progress, from target contact through manipulation and completion. Partial execution receives credit. The episode score is the sum of its subtask scores, normalized by the sum of their maxima and reported as a percentage. Completed placements remain subject to checking, so disturbing previously placed objects can reduce the score.\par
The episode terminates after 120 seconds of simulation time without an increase in the aggregate score. Only a score increase resets this shared timer; switching tasks or changing locations does not. Scores are finalized at termination. This timeout limits inactivity rather than total episode duration.\par
\begin{RAAFlow}{1. Prepare dinner}
\textbf{Setting.} The request identifies the person joining for dinner. Household memory describes their usual meal and accompaniments, while the scene shows the available food, serving equipment, and clutter in the eating area.\par
\textbf{Task.} Prepare the appropriate meal and accompaniments, serve them together on the large meal tray at the coffee table, and clear misplaced belongings and waste from the seating and serving areas.\par
\textbf{Example.} Ryan's usual dinner includes a warmed beef meal, a fork and knife, and freshly prepared coffee. The illustrated example combines meal and coffee preparation, tray serving, and clearing a book, a remote, and a used cup.\par
\par\noindent\begin{minipage}{\linewidth}
\centering
\includegraphics[width=\linewidth,height=0.62\textheight,keepaspectratio]{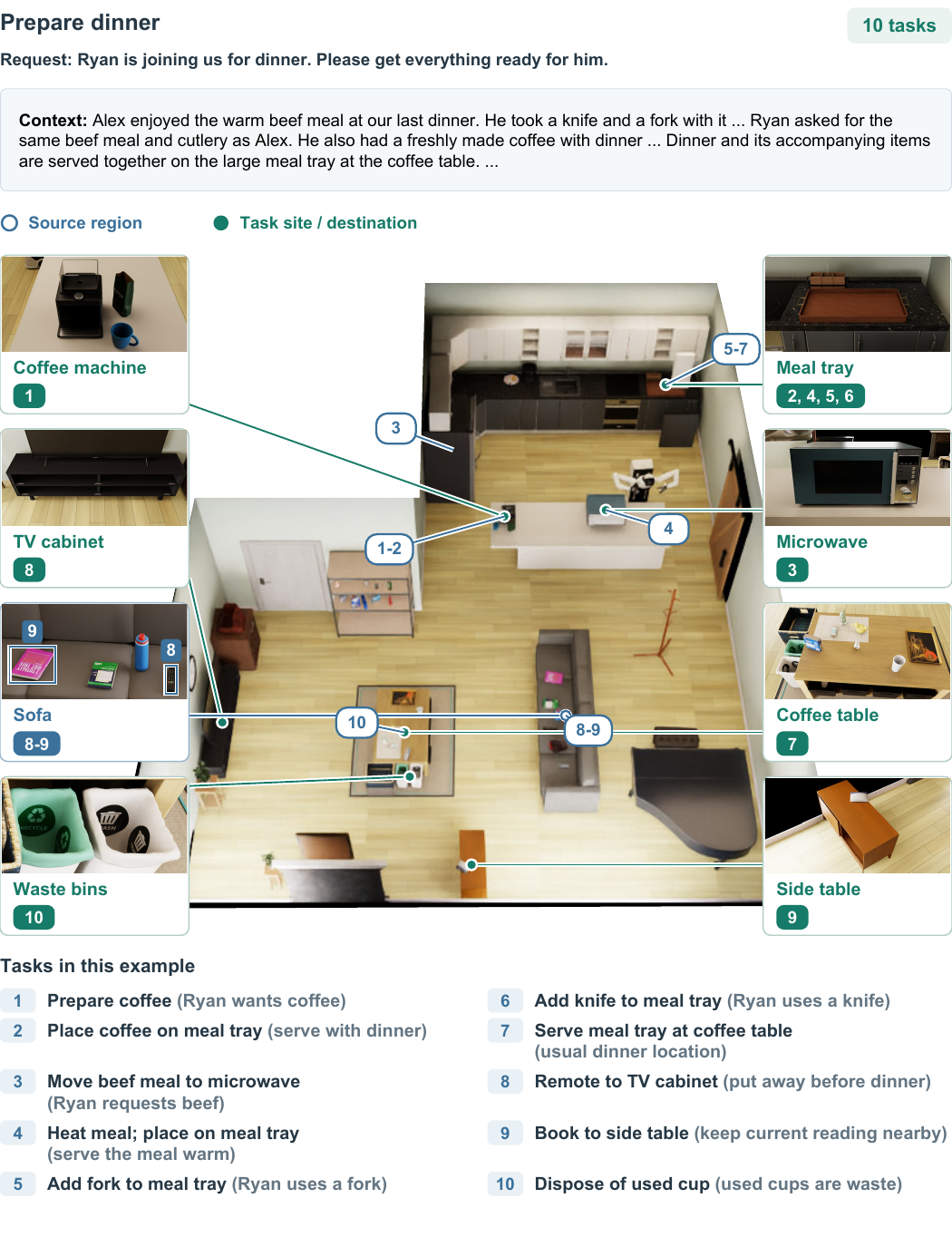}
\captionof{figure}{An example of dinner preparation with coffee.}\label{fig:openend-dinner}
\end{minipage}\par
\end{RAAFlow}

\begin{RAAFlow}{2. Host guests}
\textbf{Setting.} The request identifies the visitor but does not name the refreshments to prepare. Household memory records the visitor's usual snack and drink. Products are stocked on the shelf, and belongings or used packaging may occupy the seating and serving areas.\par
\textbf{Task.} Select the visitor's usual refreshments and serve them together on the wooden tray. Return misplaced belongings to their designated locations and dispose of waste in the appropriate bins. Other stocked products should remain on the shelf.\par
\textbf{Example.} Alex usually has Sprite and Oreos. The illustrated example combines serving these refreshments with moving the book to the side table, hanging the cap, returning the remote to the TV cabinet, storing the wallet in the basket, and disposing of a used can and cup.\par
\par\noindent\begin{minipage}{\linewidth}
\centering
\includegraphics[width=\linewidth,height=0.62\textheight,keepaspectratio]{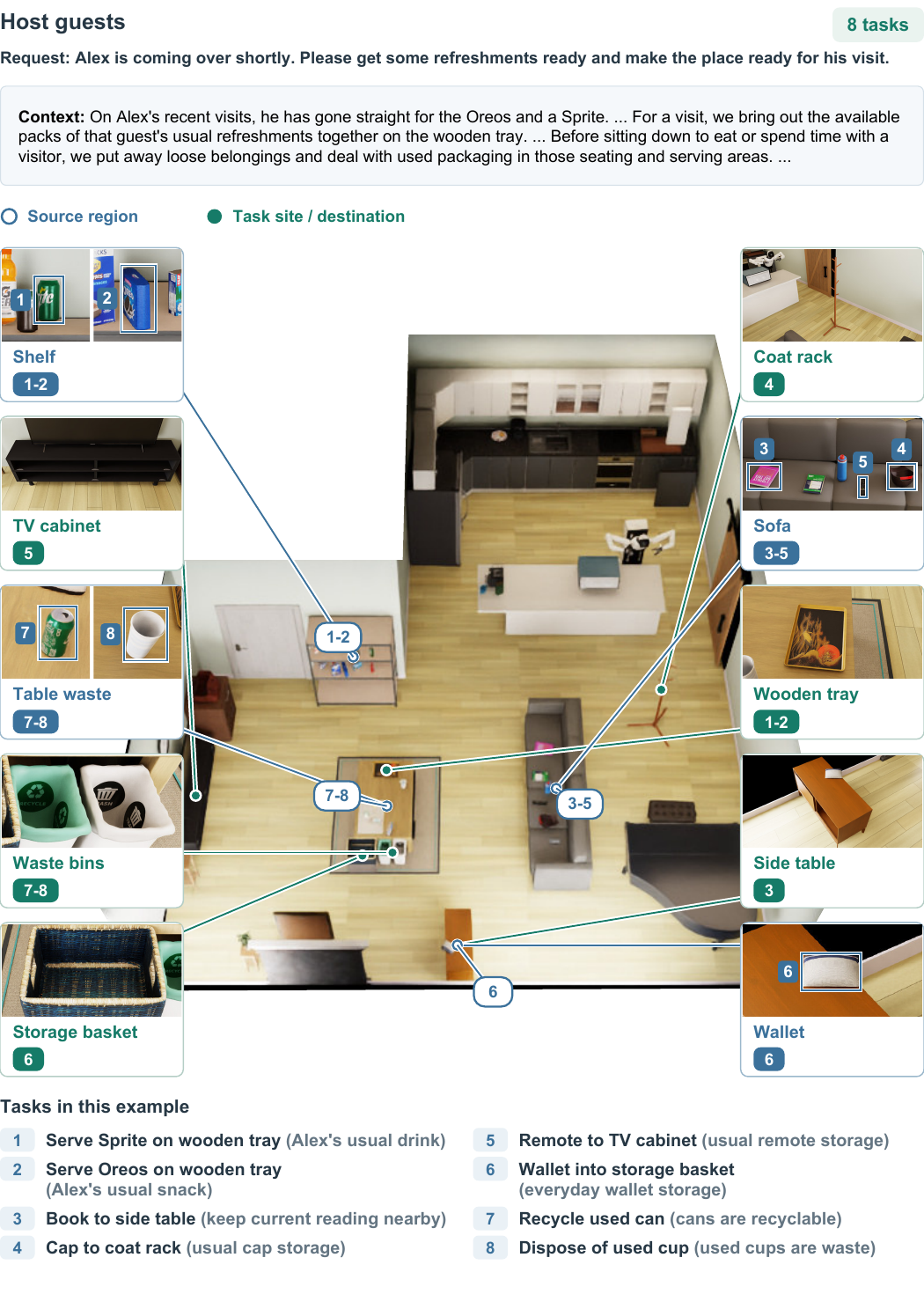}
\captionof{figure}{An example of preparing for a guest visit.}\label{fig:openend-guests}
\end{minipage}\par
\end{RAAFlow}

\begin{RAAFlow}{3. Prepare to leave}
\textbf{Setting.} The request describes who is going out and the purpose of the outing. Household memory specifies the belongings needed for that activity and where other objects belong. The scene shows which items remain unpacked or misplaced.\par
\textbf{Task.} Collect the required belongings and pack them in the open bag. Put away other misplaced objects, remove waste, and clean any table stains. Items already packed or correctly stored need no further transfer.\par
\textbf{Example.} Emma needs her wallet and power bank when working away from home. In the illustrated example, the water bottle and glasses case also need returning to their usual storage location in the open bag. The remaining tasks are to return the remote, store the notebook and close the drawer, and dispose of paper waste and a snack wrapper.\par
\par\noindent\begin{minipage}{\linewidth}
\centering
\includegraphics[width=\linewidth,height=0.62\textheight,keepaspectratio]{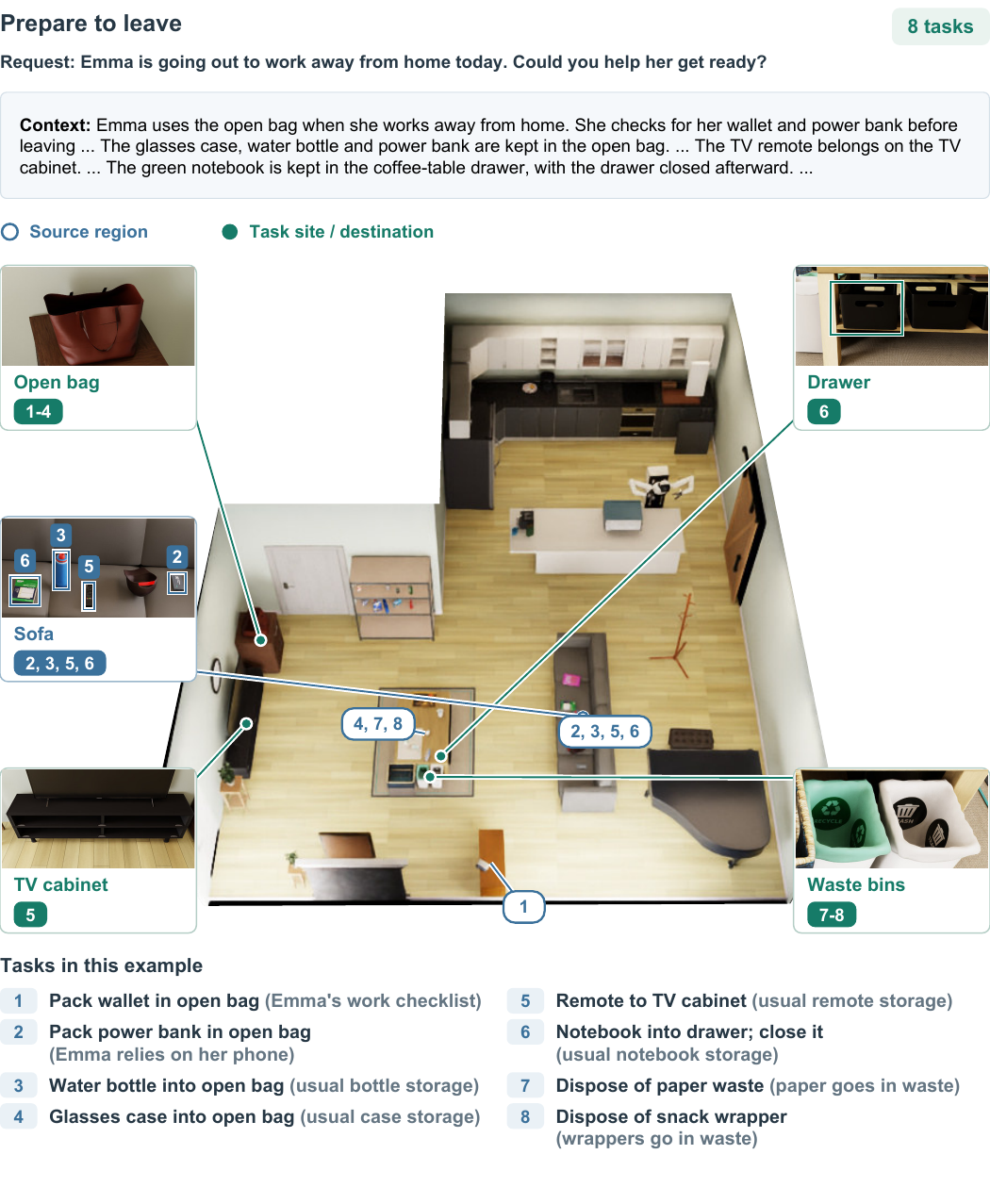}
\captionof{figure}{An example of preparing to leave.}\label{fig:openend-leave}
\end{minipage}\par
\end{RAAFlow}

\begin{RAAFlow}{4. Tidy the room}
\textbf{Setting.} Personal belongings and waste are scattered around the room, and the coffee table may have stains. Household memory specifies where belongings should be stored and how waste should be sorted.\par
\textbf{Task.} Return misplaced belongings to their designated locations, dispose of waste in the appropriate bins, and clean any table stains with the brush, returning it afterward. Objects already in their correct locations should remain undisturbed.\par
\textbf{Example.} The illustrated example includes returning a book, cap, remote, notebook, wallet, and glasses case to their designated locations, together with disposing of a used cup and recycling an empty can. No table cleaning is required in this example.\par
\par\noindent\begin{minipage}{\linewidth}
\centering
\includegraphics[width=\linewidth,height=0.62\textheight,keepaspectratio]{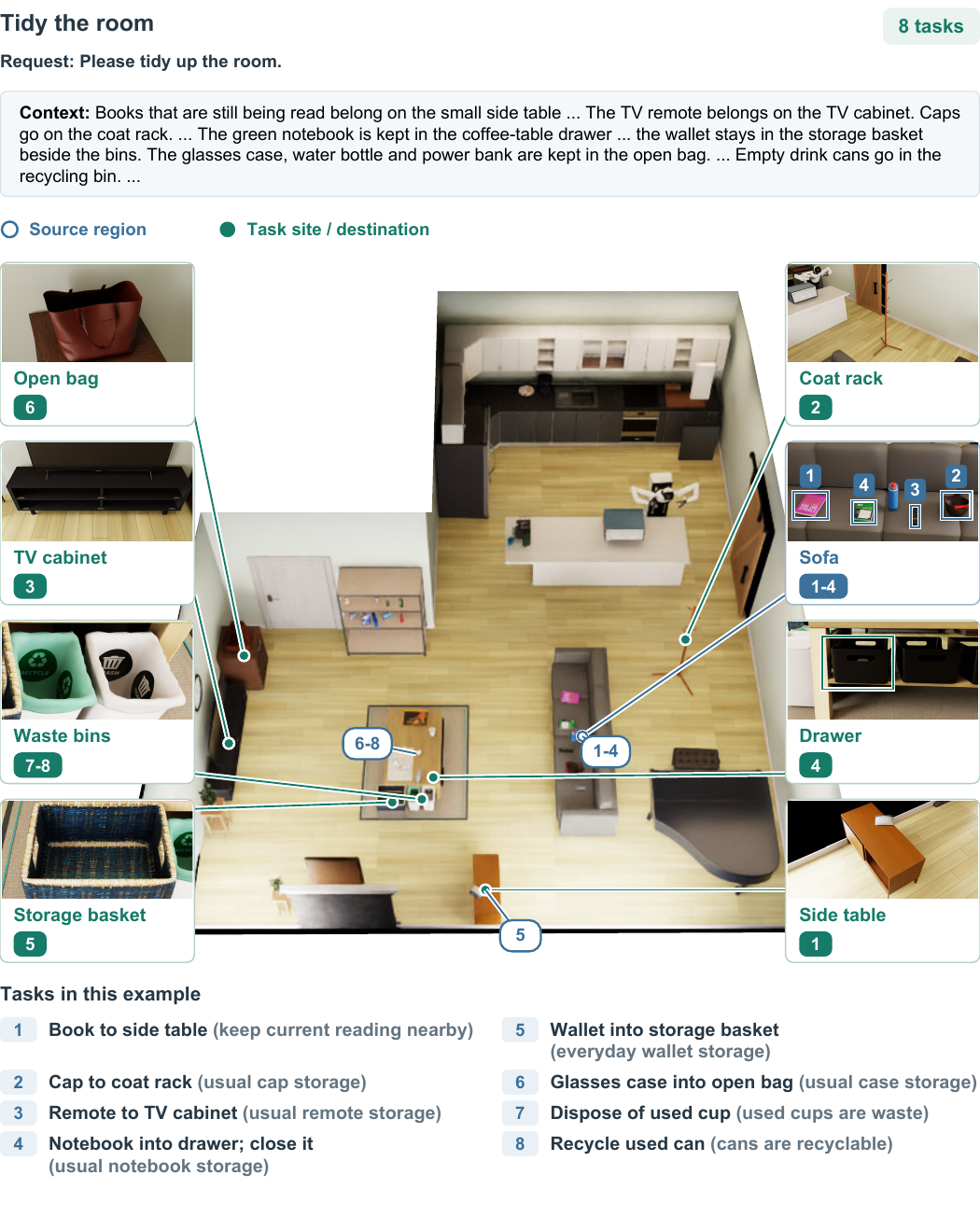}
\captionof{figure}{An example of room tidying.}\label{fig:openend-tidy}
\end{minipage}\par
\end{RAAFlow}

\par